    \documentclass[anon]{l4dc2026}
    
    \title[HGPQ]{Heteroscedastic Gaussian Process Q-learning for Sample-Efficient Experimentation}
    \usepackage{times}
    \usepackage{booktabs}
    \usepackage{array}
    \usepackage{float}
    \usepackage{placeins}
    \usepackage{wrapfig}
    \usepackage{mathtools}
    \usepackage{thm-restate}
    
    \newtheorem{assumption}{Assumption}

    \newcommand{\bx}{\mathbf{x}}
    \newcommand{\bo}{\mathbf{o}}
    \newcommand{\bu}{\mathbf{u}}
    \newcommand{\bz}{\mathbf{z}}
    \newcommand{\bw}{\mathbf{w}}
    \newcommand{\by}{\mathbf{y}}
    \newcommand{\bkappa}{\boldsymbol{\kappa}}
    
    \newcommand{\Leb}{\operatorname{Leb}}
    
    \DeclareMathOperator*{\argmax}{arg\,max}

    \coltauthor{\Name{Maximilian Bloor} \Email{max.bloor22@imperial.ac.uk}\\
     \Name{Max Mowbray} \Email{max.mowbray@imperial.ac.uk}\\
     \Name{Calvin Tsay} \Email{c.tsay@imperial.ac.uk}\\
     \Name{Antonio {del Rio Chanona}} \Email{a.del-rio-chanona@imperial.ac.uk}\\
     \addr Imperial College London, UK.}
    
\begin{document}
    
    \maketitle
    
    \begin{abstract}%
    Self-driving laboratories must learn to control a process from tens to hundreds of expensive experiments, too few for deep reinforcement learning (RL). We learn the Q-function directly with a Gaussian process (GP), which avoids learning the dynamics, and show that the GP posterior supplies the learning rate of Q-learning. If each Bellman target is treated as a noisy measurement of the Q-function, the learning rate that minimises the mean squared error is the Kalman gain, and applying it to every observed transition is equivalent to GP regression. Heteroscedastic Gaussian Process Q-learning (HGPQ) sets the noise variance of each target with a heteroscedastic GP, which learns how unreliable the targets are across the states and inputs. Targets from noisy operating conditions, or from regions where the value function has large gradients, then move the estimate less than targets from low-noise regions, so HGPQ learns quickly where the data are reliable without being misled where they are not.
     For kernel value iteration with per-target variances that are assigned when a target is collected, we prove a high-probability bound on the error of the value estimate, uniform over the explored region. Pooled over five simulated benchmarks, which include a chemical reactor and a bioprocess, HGPQ approaches the best return found by any method in fewer episodes than GP, kernel and deep RL baselines.
    \end{abstract}
    
    
    \section{Introduction}\label{sec:intro}
    
    Self-driving laboratories run experimental campaigns in chemistry and biology with automated equipment and a learning algorithm that decides what to do next \citep{abolhasani2023rise}. Each episode is a real experiment, in which the controller measures the state of a reactor or culture, sets inputs such as a feed rate, and is often rewarded only once the batch is complete \citep{mesbah_stochastic_2016, bradford_stochastic_2020}. Each campaign is therefore a finite-horizon Markov decision process (MDP) with continuous states and inputs and unknown dynamics. The controller must learn from tens to hundreds of episodes, too few to also tune the hyperparameters of the learner.
    
    Reinforcement learning (RL) with Gaussian processes (GPs) usually learns a model of the dynamics and plans through it \citep{deisenroth_gaussian_2015, bradford_stochastic_2020, bayrooti2025noregret}, but model errors compound over the horizon~\citep{janner2019trust, polymenakos2020safety}, especially when the reward arrives only at the end. We instead place a GP prior \citep{rasmussen2006gaussian} on the Q-function \citep{watkins1992q}, which the controller maximises directly to choose its inputs. Similarly, Bayesian optimisation models the objective of an experiment rather than the process behind it \citep{shahriari2016taking}.
    
    Q-learning moves its estimate towards each new Bellman target by a learning rate, which must be tuned for each problem \citep{george2006adaptive}. The noise of a laboratory process is rarely known in advance and changes between instruments, batches of material and operating conditions \citep{aldeghi2021golem, petsagkourakis2020reinforcement, griffiths2022achieving}. Kalman temporal difference (TD) methods set the rate to a Kalman gain by treating each target as a noisy measurement of the Q-function, but tune the noise variance of the targets in place of the rate \citep{george2006adaptive, geist2010kalman, tripp2013approximate}, and existing GP Q-learning methods tune the rate itself~\citep{bloor2025gaussian, chowdharyGPQ}. Variance-aware value iteration instead weights each target by the inverse of its variance, which in theory needs fewer samples \citep{min2021variance, yin2022near}. \emph{Heteroscedastic Gaussian Process Q-learning} (HGPQ) makes this practical. It estimates the variance of each target with a heteroscedastic GP and uses it in the Kalman gain, computed as GP regression. Our contributions are:
    \begin{itemize}
    \item \textbf{Method.} HGPQ, a GP Q-learning method whose learning rate is the Kalman gain and whose target variances are estimated by a heteroscedastic GP, so that no learning rate is tuned.
    \item \textbf{Theory.} A high-probability bound on the error of the posterior mean, uniform over the explored region, for per-target variances that are assigned when a target is collected. 
    \item \textbf{Results.} Pooled over five benchmarks, HGPQ approaches the best return found by any method in fewer episodes than GP, kernel and deep RL baselines with equal or larger budgets. 
    \end{itemize}

    \paragraph{Related work.} The bias-adjusted Kalman filter (BAKF) and the optimal stepsize algorithm for approximate value iteration (OSAVI) \citep{george2006adaptive, ryzhov2015new}, GP temporal differences (GPTD) and GP SARSA (GPSARSA) \citep{engel2003bayes, engel2005reinforcement} and Kalman temporal differences (KTD) for Q-learning \citep{geist2010kalman} set the TD learning rate from a single noise variance. Kalman filter Q-learning \citep{tripp2013approximate}, assumed density filtering Q-learning \citep{jeong2019assumed} and deep ensembles \citep{lee2021sunrise} give each target its own noise or weight, but from the epistemic uncertainty of the successor's value. HGPQ instead follows variance-aware value iteration \citep{min2021variance, yin2022near}. This weights each target by the inverse of an estimated variance of the next value plus a floor, and it has sharper regret and sample complexity than unweighted regression \citep{zhou2021nearly, he2023nearly, agarwal2023voql}. HGPQ estimates these variances nonparametrically with a heteroscedastic GP \citep{goldberg1998regression, kersting2007most}. Like optimistic kernel value iteration \citep{jin2020provably, yang2020provably, chowdhury2023value}, whose regret is order-optimal for Mat\'ern kernels \citep{vakili2023kernelized}, HGPQ fits a regularised estimate per time step \citep{kanagawa2018gaussian}. Like fitted Q-iteration \citep{ernst2005tree}, it regresses Bellman targets on stored transitions, and unlike GP dynamic programming \citep{deisenroth2009gaussian}, which models both the value function and the dynamics with GPs, it needs no model of the dynamics. We bound its error uniformly over the explored region of the state-input space.

    \section{Problem setting and Gaussian processes}\label{sec:prelim}
    
    \paragraph{Finite-horizon MDPs.}\label{sec:mdp} At time steps $t = 0, \ldots, T-1$ the controller observes a state $\bx_t$ in a compact set $\mathcal{X} \subset \mathbb{R}^{d_x}$ and applies an input $\bu_t$ from a box $\mathcal{U} \subset \mathbb{R}^{d_u}$. We write $\bz = (\bx, \bu)$, $\mathcal{Z} := \mathcal{X} \times \mathcal{U}$ and $d := d_x + d_u$. The pair $\bz_t$ produces a random stage reward $r_t \in [r_{\min}, r_{\max}]$ with mean $\bar r_t(\bz) := \mathbb{E}[r_t \mid \bz_t = \bz]$ and a random next state $\bx_{t+1} \sim P_t(\cdot \mid \bz_t)$, with $P_t$ the transition kernel, and the initial state is drawn from a density $p_0$. The final state $\bx_T$ produces a random terminal reward $r_T \in [r^T_{\min}, r^T_{\max}]$ with mean $\bar r_T(\bx) := \mathbb{E}[r_T \mid \bx_T = \bx]$, such as the yield of a batch. The goal is a policy $\pi = (\pi_0, \ldots, \pi_{T-1})$ that maximises the expected return $\mathbb{E}_\pi\big[\sum_{t=0}^{T-1} r_t + r_T\big]$. The optimal Q-function $Q^*_t(\bz)$, the expected return of applying $\bu$ in state $\bx$ at time step $t$ and acting optimally afterwards, satisfies the Bellman equation, solved backwards from the terminal value $V^*_T := \bar r_T$,
    \begin{equation}\label{eq:bellman}
    Q^*_t(\bz) = \bar r_t(\bz) + \mathbb{E}_{\bx' \sim P_t(\cdot \mid \bz)}\big[V^*_{t+1}(\bx')\big], \qquad V^*_t(\bx) := \max_{\bu \in \mathcal{U}} Q^*_t(\bx, \bu), \qquad t = T-1, \ldots, 0,
    \end{equation}
    with $\bar r_T \equiv 0$ without a terminal reward. The greedy policy $\pi^*_t(\bx) \in \argmax_{\bu} Q^*_t(\bx, \bu)$ is optimal, so HGPQ learns $Q^*_0, \ldots, Q^*_{T-1}$ from data and needs no model of the dynamics.
    
    \paragraph{Gaussian process regression.}\label{sec:gp} Under a GP prior $f \sim \mathcal{GP}(0, \kappa)$ on $f: \mathcal{Z} \to \mathbb{R}$ \citep{rasmussen2006gaussian}, given observations $o_i = f(\bz_i) + \eta_i$, $i = 1, \ldots, n$, with independent zero-mean noise of variance $\lambda_i$, write $\bo := (o_1, \ldots, o_n)^\top$, $\Lambda := \operatorname{diag}(\lambda_1, \ldots, \lambda_n)$, $K := [\kappa(\bz_i, \bz_j)]_{i,j \le n}$ and $\bkappa_n(\bz) := (\kappa(\bz_1, \bz), \ldots, \kappa(\bz_n, \bz))^\top$. Conditioning on $\bo$ gives the posterior $\mathcal{GP}(\mu, \Sigma)$ with
    \begin{equation}\label{eq:gp_post}
    \mu(\bz) = \bkappa_n(\bz)^\top (K + \Lambda)^{-1} \bo, \qquad
    \Sigma(\bz, \bz') = \kappa(\bz, \bz') - \bkappa_n(\bz)^\top (K + \Lambda)^{-1} \bkappa_n(\bz'),
    \end{equation}
    and we write $\sigma^2(\bz) := \Sigma(\bz, \bz)$. The noise is \emph{heteroscedastic} when the $\lambda_i$ differ. We use the Mat\'ern-5/2 kernel $\kappa(\bz, \bz') = \sigma_f^2\, k_{5/2}\big(\|\operatorname{diag}(\boldsymbol\ell)^{-1}(\bz - \bz')\|\big)$ with signal variance $\sigma_f^2$ and one lengthscale $\ell_j$ per dimension of $\bz$, except on the LMDP, where the kernel is linear in fixed features (Appendix~\ref{app:implementation}).
    
    \section{Heteroscedastic Gaussian Process Q-learning}\label{sec:method}
    
    \subsection{Bellman targets}\label{sec:targets}

    Like least-squares value iteration \citep{jin2020provably} and kernel optimistic least-squares value iteration (KOVI, \citealp{yang2020provably}), HGPQ keeps one GP per time step, $Q_t \sim \mathcal{GP}(\mu_t, \Sigma_t)$ for $t = 0, \ldots, T-1$, refitted backwards after each episode, and a dataset $\mathcal{D}_t$ of the transitions observed at time step $t$. The GPs have independent priors, because the Q-function of a finite-horizon problem differs between time steps, which are coupled only through the targets. Each stored transition gives the Bellman target, the Q-learning target with the GP mean in place of a table of values,
    \begin{equation}\label{eq:target}
    y_{t,i} = r_i + V_{t+1}(\bx'_i), \qquad V_{t+1}(\bx) := \max_{\bu \in \mathcal{U}} \mu_{t+1}(\bx, \bu),
    \end{equation}
    where $V_T$ is the terminal value. Our benchmarks give the reward of the final state on the last transition, so $V_T \equiv 0$, and only the synthetic MDPs of Section~\ref{sec:results_toy} have a known terminal reward, $V_T = \bar r_T$. Appendix~\ref{app:terminal} covers other terminal rewards and the clipping of $V_{t+1}$.

    \subsection{Learning rate from the GP posterior}\label{sec:gain}
    
    Tabular Q-learning \citep{watkins1992q} moves a stored value $\hat Q_t(\bz_i)$ towards the target of a transition $(\bz_i, r_i, \bx'_i)$ observed at time step $t$ by $\alpha\big(y_{t,i} - \hat Q_t(\bz_i)\big)$, with a learning rate $\alpha \in (0, 1]$ that must often be tuned. We instead hold a GP belief $Q_t \sim \mathcal{GP}(\mu_t, \Sigma_t)$ about the Q-function at time step $t$ and treat each target as a noisy measurement of it,
    \begin{equation}\label{eq:measurement}
    y_{t,i} = r_i + \max_{\bu \in \mathcal{U}} \mu_{t+1}(\bx'_i, \bu) = Q_t(\bz_i) + \eta_{t,i}, \qquad \mathbb{E}[\eta_{t,i}] = 0, \quad \mathrm{Var}(\eta_{t,i}) = \lambda_{t,i},
    \end{equation}
    
    with $\eta_{t,i}$ independent of $Q_t$, the model of Section~\ref{sec:gp} with $f = Q_t$. Write $\mu^{(i)}_t$ and $\Sigma^{(i)}_t$ for the belief after the first $i$ targets, starting from the prior, $\mu^{(0)}_t \equiv 0$ and $\Sigma^{(0)}_t = \kappa$. Where tabular Q-learning moves one stored value by the scalar rate $\alpha$, HGPQ moves the whole mean and covariance by the \emph{Kalman gain} $\alpha^{(i)}_t$, a function of the state--input pair $\bz$,
    \begin{equation}\label{eq:gain}
    \alpha^{(i)}_t(\bz) := \frac{\Sigma^{(i-1)}_t(\bz, \bz_i)}{\Sigma^{(i-1)}_t(\bz_i, \bz_i) + \lambda_{t,i}},
    \end{equation}
    \begin{equation}\label{eq:update}
    \begin{aligned}
    \mu^{(i)}_t(\bz) &= \mu^{(i-1)}_t(\bz) + \alpha^{(i)}_t(\bz)\big(y_{t,i} - \mu^{(i-1)}_t(\bz_i)\big),\\
    \Sigma^{(i)}_t(\bz, \bz') &= \Sigma^{(i-1)}_t(\bz, \bz') - \alpha^{(i)}_t(\bz)\,\Sigma^{(i-1)}_t(\bz_i, \bz'),
    \end{aligned}
    \end{equation}
    which minimises the mean squared error of the update at every pair (Proposition~\ref{prop:gain}, Appendix~\ref{app:proofs}).

    \begin{figure}[t]
        \centering
        \includegraphics[width=0.5\linewidth]{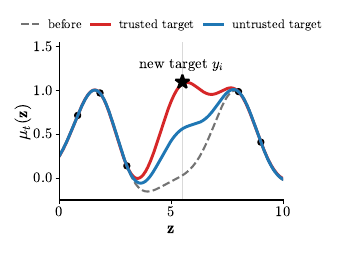}
        \vspace{-5mm}
        \caption{Posterior mean before and after one new target $y_i$, with a small (trusted) or a large (untrusted) variance $\lambda_i$. The Kalman gain moves the mean towards a trusted target and spreads the update to nearby pairs.}
        \label{fig:gain}
    \end{figure}
    Its value at the measured pair, $\alpha^{(i)}_t(\bz_i) \in [0, 1)$, plays the role of the rate $\alpha$. It is near one for an uncertain estimate and a precise target, and near zero for a precise estimate or a noisy target. It spreads the update to nearby pairs through the posterior covariance (Figure~\ref{fig:gain}). It also shrinks without a schedule as the variance falls. After all $n$ targets, in any order, the updates~\eqref{eq:update} give the GP posterior~\eqref{eq:gp_post}, written $\mu_t$ and $\Sigma_t$, with observations $\by_t := (y_{t,1}, \ldots, y_{t,n})^\top$ and noise $\Lambda_t := \operatorname{diag}(\lambda_{t,1}, \ldots, \lambda_{t,n})$ (Proposition~\ref{prop:batch}). GP Q-learning with the Kalman gain is therefore linear-Gaussian Bayesian inference, which HGPQ carries out in closed form as one GP regression. The targets of the previous time step bootstrap from $\mu_t$, and the policy acts on it.

    \subsection{Heteroscedastic target noise}\label{sec:noise}
    
    \paragraph{Decomposition of the target error.} The Kalman gain is the best learning rate only when $\lambda_{t,i}$ is the mean squared error of the target about $Q^*_t(\bz_i)$ (Proposition~\ref{prop:gain}), and the joint update of Proposition~\ref{prop:batch} treats the errors of different targets as independent. With $V^*_{t+1} := \max_{\bu} Q^*_{t+1}(\cdot, \bu)$ the optimal value, $V^*_T := \bar r_T$ and $\bar r_t$ the mean reward of Section~\ref{sec:mdp}, this error has three parts,
    \begin{equation}\label{eq:decomp}
    y_{t,i} - Q^*_t(\bz_i) = \underbrace{r_i - \bar r_t(\bz_i)}_{\text{reward}} + \underbrace{V_{t+1}(\bx'_i) - \mathbb{E}\big[V_{t+1}(\bx') \mid \bz_i\big]}_{\text{next state}} + \underbrace{\mathbb{E}\big[V_{t+1}(\bx') - V^*_{t+1}(\bx') \mid \bz_i\big]}_{\text{next estimate}}.
    \end{equation}
    The reward and the next state are drawn independently for each transition, so the first two parts are independent between targets. The third is the error of one function, $\mu_{t+1}$, averaged over nearby next states. Targets whose next states are close therefore share it, and a larger variance per target does not average it away (Appendix~\ref{app:proof_lambda}).

    \paragraph{Variance of each target.} HGPQ gives each target the variance of the part of its error that is independent between targets,
    \begin{equation}\label{eq:lambda}
    \lambda_{t,i} = \sigma_r^2 + \omega_t(\bz_i), \qquad t = T-1, \ldots, 0,
    \end{equation}
    and leaves the shared error as a bias, which vanishes as $\mu_{t+1}$ converges to $Q^*_{t+1}$. Here $\sigma_r^2$ is a small floor for the reward noise, such as the measurement noise of a yield sensor, and $\omega_t(\bz_i) \ge 0$ is the remaining independent error, which has two sources. The first is the spread of the next value over the next states reached from $\bz_i$. It varies with the pair because the process noise depends on the operating conditions \citep{griffiths2022achieving} and the gradient of the value function varies across the state space. The second, where the transitions are nearly deterministic, is the bootstrap error where $V_{t+1}$ has large gradients, such as at a step or where the maximising input switches. A small error in where $\mu_{t+1}$ places such a step gives the targets on either side errors of opposite sign.

    \paragraph{Heteroscedastic GP estimate.} HGPQ estimates $\omega_t$ from the leave-one-out residual of each target, its difference from the prediction of the GP fitted to all other targets. With $A_t := (K + \Lambda_t)^{-1}$ for the targets of time step $t$ and their current variances, the residual and its predictive variance follow in closed form from one fit \citep{rasmussen2006gaussian},
    \begin{equation}\label{eq:loo}
    e_i := y_{t,i} - \mu_{t,-i}(\bz_i) = \frac{[A_t \by_t]_i}{[A_t]_{ii}}, \qquad \sigma^2_{t,-i}(\bz_i) + \lambda_{t,i} = \frac{1}{[A_t]_{ii}}.
    \end{equation}
    An error shared by targets at nearby inputs is already predicted by the GP fitted to the other targets, so it cancels in the residual, which therefore measures the target's independent error. As in heteroscedastic GP regression \citep{goldberg1998regression, kersting2007most}, a second GP $g_t$, with its own kernel and Gaussian noise level, is fitted to the pairs $(\bz_i, \log e_i^2)$. With $m_t$ its posterior mean, shifted by the constant $-\mathbb{E}[\log\chi^2_1]$ that corrects for the logarithm (Appendix~\ref{app:fitting}),
    \begin{equation}\label{eq:omega}
    \omega_t(\bz_i) = \max\Big\{0,\ \exp\big(m_t(\bz_i)\big) - \sigma^2_{t,-i}(\bz_i) - \sigma_r^2\Big\},
    \end{equation}
    the expected squared residual less the predictive variance of the GP and the floor. The variances and the kernel are then fitted in turn.

    \subsection{Algorithm}\label{sec:algorithm}
    \begin{algorithm2e}[t]
    \DontPrintSemicolon
    \footnotesize
    \caption{Heteroscedastic Gaussian Process Q-learning (HGPQ)}\label{alg:hgpq}
    \KwIn{horizon $T$, kernel family $\kappa$, reward noise $\sigma_r^2$, exploration probability $\epsilon$, episodes $N$, and datasets $\mathcal{D}_t$ of transitions from $N_0$ episodes with space-filling inputs.}
    \For{episode $n = 1, \ldots, N$}{
        fit the terminal GP to the pairs $(\bx_T, r_T)$ in $\mathcal{D}_T$, if used (Appendix~\ref{app:terminal}) \tcp*{terminal value}
        \For(\tcp*[f]{backward pass}){$t = T-1, \ldots, 0$}{
            \For{each $(\bz_i, r_i, \bx'_i) \in \mathcal{D}_t$}{
                $y_{t,i} \leftarrow r_i + V_{t+1}(\bx'_i)$ \tcp*{target~\eqref{eq:target}}
            }
            fit the kernel by marginal likelihood and $\omega_t$ by a heteroscedastic GP \tcp*{noise model}
            $\Lambda_t \leftarrow \operatorname{diag}(\lambda_{t,i})$ with $\lambda_{t,i} = \sigma_r^2 + \omega_t(\bz_i)$ \tcp*{target variances}
            $(\mu_t, \Sigma_t) \leftarrow$ posterior~\eqref{eq:gp_post} of $Q_t$ given $\by_t$ with noise $\Lambda_t$ \tcp*{\eqref{eq:update} on every target}
        }
        \For(\tcp*[f]{rollout}){$t = 0, \ldots, T-1$}{
            draw $\tilde Q_t \sim \mathcal{GP}(\mu_t, \Sigma_t)$ \tcp*{Thompson sampling}
            $\bu_t \leftarrow \argmax_{\bu} \tilde Q_t(\bx_t, \bu)$, or uniform on $\mathcal{U}$ with probability $\epsilon$\;
            Apply $\bu_t$, observe $r_t$ and $\bx_{t+1}$, and add $(\bz_t, r_t, \bx_{t+1})$ to $\mathcal{D}_t$\;
        }
        observe the terminal reward $r_T$ of $\bx_T$, if any, and add $(\bx_T, r_T)$ to $\mathcal{D}_T$\;
    }
    \KwOut{after a final backward pass, the greedy policy $\hat\pi_t(\bx) \in \argmax_{\bu} \mu_t(\bx, \bu)$.}
    \end{algorithm2e}
    
    Algorithm~\ref{alg:hgpq} summarises HGPQ. The targets of time step $t$ depend on $\mu_{t+1}$, so each backward pass fits the GPs from the terminal value $V_T$ down to $t = 0$, as the Bellman equation~\eqref{eq:bellman} is solved,
    \begin{equation}\label{eq:chain}
    \mu_{t+1} \xrightarrow[\text{\eqref{eq:target}, \eqref{eq:lambda}}]{\text{targets, variances}} (\by_t, \Lambda_t) \xrightarrow[\text{\eqref{eq:gp_post}}]{\text{GP fit}} (\mu_t, \Sigma_t), \qquad t = T-1, \ldots, 0.
    \end{equation}
    Each posterior supplies the targets and variances of the time step before it, with $\mu_{t+1}$ held fixed, and after each episode the whole chain is refitted from the prior, since a change in $\mu_{t+1}$ changes every target of time step $t$. After $N_0$ initial episodes with space-filling inputs, the behaviour policy applies a uniformly random input with probability $\epsilon$, which maintains coverage, and otherwise maximises a posterior sample $\tilde Q_t$ \citep{thompson1933likelihood} drawn by pathwise conditioning \citep{wilson2020efficiently, wilson2021pathwise} (Appendix~\ref{app:implementation}).
    
\section{Convergence in function space}\label{sec:analysis}\label{sec:convergence}

The greedy policy evaluates $\mu_t$ at pairs never visited, so $\mu_t$ must be accurate over a whole region. Write $\mu^n_t$ for the mean after $n$ episodes. The analysis takes rewards in $[0,1]$, exact maximisation, a fixed kernel, unstandardised targets and $N_0=0$, and gives each target a base variance $\lambda_{t,i}\in[\lambda_-,\lambda_+]$, $\lambda_->0$, fixed before the outcome of its transition is drawn and multiplied after $n$ episodes by $n^{\zeta}$, $\zeta\in[0,1)$, with $\zeta=0$ in Algorithm~\ref{alg:hgpq}. It assumes (Assumption~\ref{ass:main}, Appendix~\ref{app:convergence}) Markov transitions, Bellman completeness in the reproducing kernel Hilbert space of the kernel, as in kernel value iteration \citep{yang2020provably,jin2020provably}, a kernel approximation property with exponent $q$ and constant $L_\kappa\ge0$ ($q=1$ and $L_\kappa>0$ for the Mat\'ern-5/2 kernel), and visited pairs with a density bounded below on regions $\mathcal{Z}_t:=\mathcal{X}_t\times\mathcal{U}$ whose next states stay in $\mathcal{X}_{t+1}$.

\begin{theorem}[Uniform convergence]\label{thm:main}
Let Assumption~\ref{ass:main} hold, let $\zeta\in[0,1)$ and $\delta\in(0,1)$, and let $\theta:=q/(2q+d)$ if $L_\kappa>0$ and $\theta:=1/2$ if $L_\kappa=0$. There are constants $C_{\sigma,t}$, a burn-in $n_0(\delta)$ and widths $\beta_{t,n}(\delta)$ (Definitions~\ref{def:rate} and~\ref{def:width}) such that, with probability at least $1-2\delta$, for all $n\ge n_0(\delta)$ and all time steps $t$,
\begin{equation}\label{eq:main}
\sup_{\mathbf{z}\in\mathcal{Z}_t}\bigl|\mu^n_t(\mathbf{z})-Q^*_t(\mathbf{z})\bigr|\le n^{-(1-\zeta)\theta}\sum_{j=t}^{T-1}C_{\sigma,j}\,\beta_{j,n}(\delta).
\end{equation}
If $n^{-(1-\zeta)\theta}\beta_{j,n}(\delta)\to0$ for every $j$ and every $\delta\in(0,1)$, then $\mu^n_t\to Q^*_t$ uniformly on $\mathcal{Z}_t$ almost surely.
\end{theorem}

Here the constant $C_{\sigma,t}$ grows with the largest variance $\lambda_+$ and with the inverse of the density of the visited pairs, the burn-in $n_0(\delta)$ is the number of episodes after which the visited pairs cover every region $\mathcal{Z}_t$ at the resolution of the bound, and the width $\beta_{t,n}(\delta)$ adds the norm bound of the completeness assumption, a discretisation term and a noise term that grows with the information gain of the kernel.
The proof is in Appendix~\ref{app:convergence}. It covers variances assigned when a target is collected and a fixed kernel, not the re-estimated variances or the refitted kernel of Algorithm~\ref{alg:hgpq}. For the Mat\'ern-5/2 kernel with uniformly bounded eigenfunctions, $q=1$ and $\theta=1/(2+d)$, and without a schedule the error is of order $n^{-(10-d^2)/(2(2+d)(5+d))}$ up to logarithmic factors, which tends to zero for $d\le3$. With $\zeta=d/(5+2d)$ it is of order $n^{-(5+d)/((2+d)(5+2d))}$ in every dimension (Corollary~\ref{cor:rates}). The burn-in grows at least exponentially with $d$.

    \section{Experiments}\label{sec:experiments}
    \begin{figure}[t]
        \centering
        \includegraphics[width=0.85\linewidth]{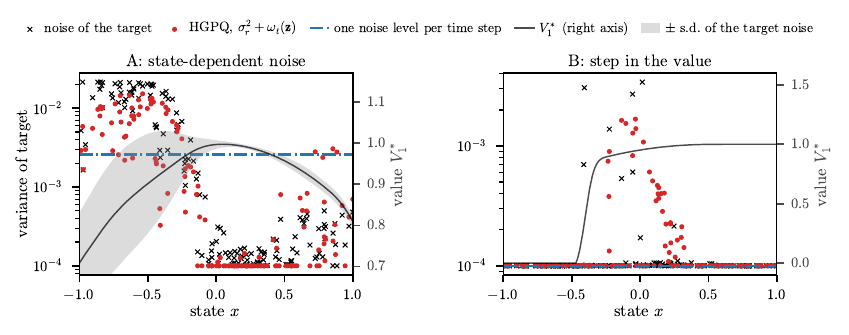}
        \vspace{-2mm}
        \caption{Two synthetic MDPs with a scalar state (Section~\ref{sec:results_toy}), one dataset each. In A the transitions from states below about $-0.2$ are noisy, and in B they are nearly deterministic and the terminal reward is a step. HGPQ's variance of each target of time step 1 and the noise of that target, against the state of its transition, one noise level per time step fitted by marginal likelihood, and the optimal value $V^*_1$ (right axis) with, in A, one standard deviation of the target noise about it. The floor in these MDPs is $\sigma_r^2 = 10^{-4}$ (Appendix~\ref{app:toy}), the lower edge of the variance axes.}
        \label{fig:toy}
    \end{figure}
    \begin{figure}[t]
        \centering
        \includegraphics[width=0.5\linewidth]{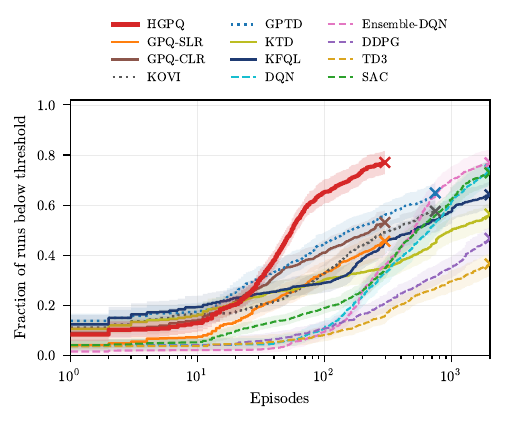}
        \caption{Fraction of runs whose normalised regret is below each of 15 thresholds from 0.5 to 0.01, pooled over the benchmarks (30 seeds, 95\% bands, crosses at the end of a budget).}
        \label{fig:ecdf}
    \end{figure}
    \begin{figure}[t]
        \centering
        \includegraphics[width=\linewidth]{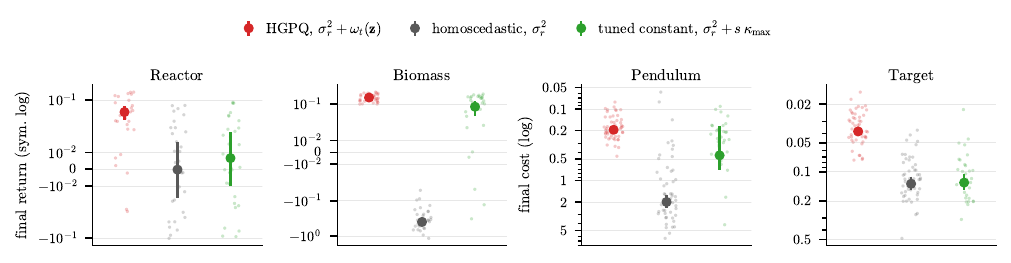}
        \vspace{-4mm}
        \caption{Final return after 300 episodes of HGPQ, of the floor $\sigma_r^2$ alone and of a constant variance $\sigma_r^2 + s\,\kappa_{\max}$ with the scale $s$ tuned per benchmark and $\kappa_{\max}$ the prior variance of the kernel (mean, 95\% interval and seeds, 30 seeds, 50 on the pendulum and target tracking). Returns of the reactor and biomass production are on symmetric logarithmic axes, and costs of the pendulum and target tracking on inverted logarithmic axes.}
        \label{fig:factorial}
    \end{figure}
    \begin{table}[t]
    \centering

    \caption{Update rules at time step $t$, with $y_i$ the Bellman target of transition $i$. GPQ-CLR moves a stored value $\hat Q_i$ towards $y_i$ with a tuned rate $\alpha$ and fits a GP to these values, KTD filters the weights $\mathbf{w}$ of fixed features $\boldsymbol\phi_i$ with Kalman gain $\mathbf{k}$, GPTD observes each reward as a difference of successive values plus noise $\eta_t$, and KOVI adds $c$ posterior standard deviations of the next GP to its target. All but GPQ-CLR move the estimate by a Kalman gain and differ in what they observe and with what noise.}
\label{tab:update_rules}
\footnotesize
\renewcommand{\arraystretch}{1.05}
\begin{tabular}{@{}>{\raggedright\arraybackslash}p{3.6cm}>{\raggedright\arraybackslash}p{6.4cm}>{\raggedright\arraybackslash}p{3.7cm}@{}}
\toprule
\textbf{Method} & \textbf{Update} & \textbf{Learning rate} \\
\midrule
GPQ-CLR {\footnotesize\citep{bloor2025gaussian}} & $\hat Q_i \leftarrow \hat Q_i + \alpha\,(y_i - \hat Q_i)$, GP fit to the $\hat Q_i$ & tuned $\alpha$ \\
KTD {\footnotesize\citep{geist2010kalman}} & $\mathbf{w} \leftarrow \mathbf{w} + \mathbf{k}\,(y_i - \boldsymbol\phi_i^\top \mathbf{w})$ & Kalman gain on fixed features, one noise level \\
GPTD {\footnotesize\citep{engel2005reinforcement}} & one GP conditioned on $r_t = Q(\bz_t) - Q(\bz_{t+1}) + \eta_t$ & Kalman gain, one noise level \\
KOVI {\footnotesize\citep{yang2020provably}} & $\mu_t = \bkappa^\top (K + \lambda I)^{-1} \by$, $y_i = r_i + \max_{\bu} (\mu_{t+1} + c\,\sigma_{t+1})$ & Kalman gain, one noise level \\
\textbf{HGPQ (ours)} & $\mu_t = \bkappa^\top (K + \Lambda)^{-1} \by$, $\Lambda_{ii} = \sigma_r^2 + \omega_t(\bz_i)$ & Kalman gain, one noise level per target \\
\bottomrule
\end{tabular}
    \end{table}

    \subsection{Experimental setup}\label{sec:setup}

    \paragraph{Benchmarks.} The \emph{non-stationary linear MDP} (LMDP) has a reward whose input coefficient changes sign every two time steps. The \emph{semi-batch reactor} \citep{bradford_economic_2018, bloor2025pc} and \emph{photosynthetic biomass production} \citep{bloor2025pc, bradford_stochastic_2020} are nonlinear processes rewarded at the end of the batch and penalised whenever a state constraint is violated, with rate constants that fluctuate between time steps, which imitates unmodelled changes in mass transfer and mixing. The \emph{inverted pendulum} is swung up and rewarded for its final state, and in \emph{target tracking} the final distance of a spring-mass system to a target is penalised. All have random initial states and process noise, and horizons of 8 to 20 time steps (Appendix~\ref{app:models}).
    
    \paragraph{Baselines and protocol.} We compare HGPQ with GP Q-learning with a scheduled or a constant learning rate (GPQ-SLR and GPQ-CLR, \citealp{bloor2025gaussian}), with KOVI \citep{yang2020provably}, with GPTD in its SARSA form \citep{engel2005reinforcement}, with KTD, with Kalman filter Q-learning (KFQL, \citealp{tripp2013approximate}), and with DQN \citep{mnih2015human}, Ensemble-DQN \citep{osband2016deep}, DDPG \citep{lillicrap_continuous_2019}, TD3 \citep{fujimoto_addressing_2018} and SAC \citep{haarnoja2018soft}. Table~\ref{tab:update_rules} compares their updates with HGPQ's. All methods start from the same $N_0$ initial episodes (5 on the LMDP, 10 on the pendulum and target tracking and 20 on the reactor and biomass production), which the budgets below do not count, are tuned on ten separate seeds and run on thirty seeds, or fifty for the comparison of target variances on the pendulum and target tracking (Appendix~\ref{app:protocol}). The budgets favour the kernel and deep RL baselines: 300 episodes for HGPQ and the other GP Q-learning methods, 750 for KOVI and GPTD, and 2000 for KTD, KFQL and deep RL. HGPQ stops at 300 episodes because an exact refit costs $O(n^3)$ per time step, and Sparse HGPQ (Appendix~\ref{app:scaling}) runs for 600 episodes at linear cost.
    
    \paragraph{Measuring sample efficiency.} The normalised regret is the gap to the best return that any method found, rather than to the true optimum, divided by the gap between that return and the starting return. As in optimiser benchmarking \citep{hansen2021coco}, a run reaches each of 15 thresholds at the first episode at which its normalised regret, averaged over the last ten episodes, falls below it.

    \subsection{Target variances in two synthetic MDPs}\label{sec:results_toy}

    Two synthetic MDPs with a scalar state and 150 random transitions per time step separate the two kinds of target error that $\omega_t$ estimates (Figure~\ref{fig:toy}, Appendix~\ref{app:toy}). In A, the transitions from states below about $-0.2$ are noisy. HGPQ's variances follow the noise of each target, whereas one noise level fitted by marginal likelihood is too small for the noisy targets and too large for the others. Where the transitions are almost noise-free ($x \ge 0$), HGPQ's value error is therefore half that with one noise level and about 25 times smaller than with the floor alone (Table~\ref{tab:toy}). In B, the transitions are nearly deterministic and a step in the terminal reward is carried back as a sharp rise in $V^*_t$. A small error in where $\mu_{t+1}$ places the step gives the targets on either side errors of opposite sign. HGPQ gives them larger variances, whereas one noise level collapses to the floor, and for $x \ge 0$, beyond the step, its value error is less than a quarter of that with one noise level, at the price of a larger error for $x < 0$, which contains the step. Its greedy policy is closer to the optimal return in B and about as close in A (Appendix~\ref{app:toy_results}).

    \subsection{Sample efficiency}\label{sec:results_efficiency}

    Pooled over the benchmarks, HGPQ comes closest to the best returns in the fewest episodes (Figure~\ref{fig:ecdf}). It reaches 69\% of the thresholds within 150 episodes and 77\% within 300, more than every baseline after its full budget of 750 or 2000 episodes except Ensemble-DQN, which reaches as many after 2000. GPQ-SLR and GPQ-CLR, which update a table of stored values with a scheduled or tuned learning rate, reach about half the thresholds by the end of their 300-episode budget. Per benchmark, HGPQ is the fastest on the pendulum and target tracking. On the reactor it reaches more thresholds within 300 episodes than any baseline within the same budget, and only DQN, Ensemble-DQN and KFQL reach more, after 2000 episodes. On biomass production GPQ-CLR matches HGPQ within 150 episodes, and only SAC reaches more, after 2000. On the LMDP, whose value is linear, KOVI, KTD and KFQL do about as well as HGPQ (Appendix~\ref{app:extra_results}). Sparse HGPQ, which uses sparse GPs with 256 inducing points so that a refit costs time linear in $n$, remains the fastest method over 600 episodes on the pendulum and target tracking with up to ten times the process noise, more pendulums or longer horizons, in all but one setting (Appendix~\ref{app:scaling}).

    \subsection{Per-target and constant variances}\label{sec:results_parts}

    Figure~\ref{fig:factorial} compares HGPQ's per-target variances with two homoscedastic choices, with its GP, exploration and settings unchanged. The floor $\sigma_r^2$ alone, a homoscedastic GP regression with too little noise, is worse on all four benchmarks, and so is a constant variance $\sigma_r^2 + s\,\kappa_{\max}$, with $\kappa_{\max}$ the prior variance of the kernel and the scale $s$ tuned per benchmark. On the pendulum and target tracking, which are nearly deterministic and whose values have large gradients, most of the error of the targets is bootstrap error, as in MDP B. On the reactor and biomass production the fluctuating reaction and growth rates make the noise depend on the state, as in MDP A, and HGPQ's final return is well above that of the tuned constant, 0.059 against 0.007 on the reactor. HGPQ shares only the floor $\sigma_r^2$ with the constant, and a single floor $\sigma_r = 10^{-3}$ does not significantly lower the final return on any benchmark tested (Appendix~\ref{app:floor}).

    Adding the next GP's variance $\sigma^2_{t+1}$, which is large while the next estimate is poor, to every target's variance slows early learning on the pendulum and target tracking and changes the final return by at most 5\%, except for a lower final cost on target tracking. A tuned constant plus $\sigma^2_{t+1}$ matches HGPQ's final return only on these two benchmarks (Appendix~\ref{app:ablation_next}).

    \section{Conclusion}\label{sec:conclusion}
    
    We show that GP Q-learning needs no tuned learning rate. The optimal rate is the Kalman gain, which one GP regression applies to every stored transition. HGPQ gives each Bellman target the variance of the part of its error that is independent between targets, estimated by a heteroscedastic GP: the noise of the process where it is noisy, and the local bootstrap error. Pooled over five benchmarks, it needs fewer episodes than GP, kernel and deep RL baselines, and on the reactor and biomass benchmarks, where the noise depends on the state, it outperforms a constant variance tuned per benchmark. The analysis assumes a fixed kernel and variances set when a target is collected. The $O(n^3)$ cost of an exact refit per time step is small beside an experiment lasting hours, and Sparse HGPQ makes it linear in $n$. The benchmarks penalise violations of the state constraints rather than prevent them, and future work will handle the constraints with feasibility backups.
    
    \bibliography{references}
    
    \appendix
    
    \section{Proofs}\label{app:proofs}

    \subsection{Optimality of the Kalman gain}\label{app:proof_gain}

    The next result concerns a single update from a belief $\mathcal{GP}(\mu, \Sigma)$. With $\mu = \mu^{(i-1)}_t$, $\Sigma = \Sigma^{(i-1)}_t$, $y = y_{t,i}$ and $\lambda = \lambda_{t,i}$, its learning rate is the gain~\eqref{eq:gain} and its update is~\eqref{eq:update}.

    \begin{proposition}[Kalman gain]\label{prop:gain}
    Let $Q \sim \mathcal{GP}(\mu, \Sigma)$ and $y = Q(\bz_i) + \eta$, with $\eta$ independent of $Q$, $\mathbb{E}[\eta] = 0$ and $\mathrm{Var}(\eta) = \lambda > 0$.
    \begin{enumerate}
    \item[(a)] For every $\bz \in \mathcal{Z}$, among the estimates $\mu(\bz) + a\,(y - \mu(\bz_i))$ with $a \in \mathbb{R}$, the mean squared error $\mathbb{E}\big[(Q(\bz) - \mu(\bz) - a\,(y - \mu(\bz_i)))^2\big]$ is smallest for the learning rate
    \[
    a = \alpha(\bz; \bz_i) := \frac{\Sigma(\bz, \bz_i)}{\sigma^2(\bz_i) + \lambda}.
    \]
    The updated estimate $\mu^+(\bz) := \mu(\bz) + \alpha(\bz; \bz_i)\,(y - \mu(\bz_i))$ has error covariance $\Sigma^+(\bz, \bz') = \Sigma(\bz, \bz') - \alpha(\bz; \bz_i)\,\Sigma(\bz_i, \bz')$. If $Q$ and $\eta$ are Gaussian, $\mu^+$ and $\Sigma^+$ are the exact posterior mean and covariance of $Q$ given $y$.
    \item[(b)] At the measured pair the learning rate is $\alpha(\bz_i; \bz_i) = \sigma^2(\bz_i) / (\sigma^2(\bz_i) + \lambda) \in [0, 1)$. If the same pair is measured $n$ times, starting from variance $\bar\sigma^2 > 0$ there, the learning rate of the $n$-th update is
    \begin{equation}\label{eq:one_over_n}
    \alpha_n = \frac{1}{n + \lambda / \bar\sigma^2} .
    \end{equation}
    \end{enumerate}
    \end{proposition}

    \noindent\emph{Proof.} \emph{Part (a).} Let $e := y - \mu(\bz_i)$. Since $\eta$ is independent of $Q$ with mean zero, $\mathrm{Var}(e) = \sigma^2(\bz_i) + \lambda$ and $\mathrm{Cov}(Q(\bz), e) = \Sigma(\bz, \bz_i)$, so the mean squared error is the parabola
    \[
    \sigma^2(\bz) - 2a\,\Sigma(\bz, \bz_i) + a^2\big(\sigma^2(\bz_i) + \lambda\big)
    \]
    in $a$, whose unique minimiser is $\alpha(\bz; \bz_i)$. With $c := \Sigma(\bz, \bz_i)\Sigma(\bz_i, \bz')/(\sigma^2(\bz_i) + \lambda)$, expanding the covariance of $Q(\bz) - \mu^+(\bz) = Q(\bz) - \mu(\bz) - \alpha(\bz; \bz_i)\,e$ and $Q(\bz') - \mu^+(\bz')$ gives $\Sigma(\bz, \bz') - 2c + c = \Sigma^+(\bz, \bz')$, since each cross term equals $-c$ and the last term equals $c$. In the Gaussian case $(Q(\bz), Q(\bz'), y)$ is jointly Gaussian, and conditioning on $y$ gives $\mu^+$ and $\Sigma^+$.

    \emph{Part (b).} The gain at $\bz_i$ lies in $[0, 1)$ because $\lambda > 0$. By part (a) the variance at the measured pair becomes $\sigma^2(\bz_i)\lambda/(\sigma^2(\bz_i) + \lambda)$, so each measurement adds $1/\lambda$ to the precision there. The variance before the $n$-th update is therefore $\bar\sigma^2_{n-1} = (1/\bar\sigma^2 + (n-1)/\lambda)^{-1}$, and $\alpha_n = 1/(1 + \lambda/\bar\sigma^2_{n-1}) = 1/(n + \lambda/\bar\sigma^2)$. \jmlrQED

    \subsection{Kalman updates as linear-Gaussian Bayesian inference}\label{app:proof_batch}

    Applying Proposition~\ref{prop:gain} to the stored transitions one at a time is a Kalman filter with a constant state, the Q-function, and hence recursive Bayesian regression \citep{geist2010kalman}.

    \begin{proposition}[Kalman updates are linear-Gaussian Bayesian inference]\label{prop:batch}
    Applying the update of Proposition~\ref{prop:gain}(a), starting from $\mathcal{GP}(0, \kappa)$, once for each of $(\bz_i, y_i, \lambda_i)$, $i = 1, \ldots, n$, in any order, gives the posterior of the model $y_i = Q(\bz_i) + \eta_i$ with independent $\eta_i \sim \mathcal{N}(0, \lambda_i)$, namely~\eqref{eq:gp_post} with $\bo = \by$ and $\Lambda = \operatorname{diag}(\lambda_1, \ldots, \lambda_n)$. The result is the same for any noise distribution with these variances.
    \end{proposition}

    \noindent\emph{Proof.} The updates and~\eqref{eq:gp_post} are algebraic in the inputs, targets, variances and kernel, so it suffices to take Gaussian noise. Then each update is exact conditioning on one target (Proposition~\ref{prop:gain}(a)), and since the noise terms are independent given $Q$, conditioning one target at a time in any order equals conditioning on $\by$ at once. With $\mathrm{Cov}(\by) = K + \Lambda$ and $\mathrm{Cov}(Q(\bz), \by) = \bkappa_n(\bz)^\top$, this gives~\eqref{eq:gp_post}. \jmlrQED

    \subsection{Derivation of the target variances}\label{app:proof_lambda}

    The measurement model~\eqref{eq:measurement} treats the error of each target,
    \[
    \begin{aligned}
        \eta_{t,i} &\coloneq y_{t,i} - Q^*_t(\bz_i) \\
        &= \underbrace{r_i - \bar r_t(\bz_i)}_{a_i} + \underbrace{V_{t+1}(\bx'_i) - \mathbb{E}\big[V_{t+1}(\bx') \mid \bz_i\big]}_{b_i} + \underbrace{\mathbb{E}\big[V_{t+1}(\bx') - V^*_{t+1}(\bx') \mid \bz_i\big]}_{c_i},
    \end{aligned}
    \]
    as noise with mean zero, and Proposition~\ref{prop:batch} treats the errors of different targets as independent. For $t < T-1$ and under the following assumption, the mean is approximately zero and the errors of the targets of a time step have covariance $D_t + S_t$, with $D_t$ diagonal, from the reward and the next state, and $S_t$ a kernel, from the error of the next estimate. HGPQ's variances~\eqref{eq:lambda} estimate $D_t$. This derivation concerns the choice of variances only and is not needed for Theorem~\ref{thm:main}, which holds for any variances that satisfy Assumption~\ref{ass:main}(v).
    \begin{assumption}[Target variances]\label{ass:lambda}
    At a time step $t < T-1$, the following hold for every stored transition $(\bz_i, r_i, \bx'_i)$.
    \begin{enumerate}
    \item[(i)] Given the data of later time steps, $\mu_{t+1}$ is fixed and $Q^*_{t+1} = \mu_{t+1} + \varepsilon$ with $\varepsilon \sim \mathcal{GP}(0, \Sigma_{t+1})$, as if the optimal Q-function were drawn from the posterior.
    \item[(ii)] Given $\bz_i$, the error $\varepsilon$ is independent of the process draw $(r_i, \bx'_i)$, and $r_i$ is uncorrelated with $V_{t+1}(\bx'_i)$.
    \item[(iii)] For every next state $\bx'$, $\mu_{t+1}(\bx', \cdot)$ has a unique maximiser $\bu^*(\bx')$ on $\mathcal{U}$ with quadratic growth, $\mu_{t+1}(\bx', \bu^*(\bx')) - \mu_{t+1}(\bx', \bu) \ge \varrho\, \|\bu - \bu^*(\bx')\|^2$ for all $\bu \in \mathcal{U}$ and some $\varrho > 0$, and $\varepsilon(\bx', \cdot)$ is Lipschitz on $\mathcal{U}$ with constant $G(\bx')$.
    \item[(iv)] The clipping in~\eqref{eq:target} is inactive, so $V_{t+1}(\bx') = \max_{\bu} \mu_{t+1}(\bx', \bu)$.
    \end{enumerate}
    \end{assumption}

    \paragraph{Mean of the error.} The reward and next-state parts have mean zero by definition, and the error of the next estimate has mean zero to first order. Expectations are over the process draw of $(r_i, \bx'_i)$ given $\bz_i$ and over $\varepsilon$. By the definitions of $\bar r_t$ and $P_t$, $\mathbb{E}[a_i] = \mathbb{E}[b_i] = 0$. For $c_i$, fix $\bx'$, write $\bu^* := \bu^*(\bx')$ and let
    \[
    R(\bx') := \max_{\bu} \big(\mu_{t+1}(\bx', \bu) - \mu_{t+1}(\bx', \bu^*) + \varepsilon(\bx', \bu) - \varepsilon(\bx', \bu^*)\big)
    \]
    be the amount by which the maximum of $\mu_{t+1} + \varepsilon$ exceeds its value at $\bu^*$. By (i) and (iv), $V^*_{t+1}(\bx') - V_{t+1}(\bx') = \varepsilon(\bx', \bu^*) + R(\bx')$, and $0 \le R(\bx') \le G(\bx')^2/(4\varrho)$. The lower bound follows from $\bu = \bu^*$, and the upper bound from (iii), since each term of the maximum is at most $G(\bx')\, s - \varrho\, s^2$ with $s := \|\bu - \bu^*\|$. The remainder is quadratic in $\varepsilon$, so to first order
    \begin{equation}\label{eq:c_first_order}
    c_i \approx -\mathbb{E}\big[\varepsilon(\bx', \bu^*(\bx')) \mid \bz_i\big], \qquad \mathbb{E}[c_i] \approx 0,
    \end{equation}
    and $\mathbb{E}[\eta_{t,i}] \approx 0$. Since $R \ge 0$, the neglected remainder makes the target underestimate the optimal value on average, by at most $\mathbb{E}\big[G(\bx')^2\big]/(4\varrho)$. Condition (iii) fails when the maximiser lies at a corner of a bang-bang input set, as on the LMDP with inputs $\{-1, 1\}$, where the expansion~\eqref{eq:c_first_order} does not apply.

    \paragraph{Covariance of the errors.} The reward and next-state parts are independent between targets, and the error of the next estimate is shared by targets whose next states are close. Let $i$ and $j$ be transitions of time step $t$, with $i = j$ allowed. By Assumption~\ref{ass:lambda}(ii), $c_j$ depends on $\varepsilon$ alone, which is independent of the process draws, and $a_i + b_i$ has mean zero given the inputs, so $\mathrm{Cov}(a_i + b_i, c_j) = 0$. The process draws of different transitions are independent given their inputs, so $\mathrm{Cov}(a_i + b_i, a_j + b_j)$ vanishes for $i \ne j$ and equals $\mathrm{Var}(r_i \mid \bz_i) + \mathrm{Var}\big(V_{t+1}(\bx') \mid \bz_i\big)$ for $i = j$, by (ii). The terms $c_i$ and $c_j$ are averages of the same error $\varepsilon$, and by~\eqref{eq:c_first_order}
    \[
        \mathrm{Cov}(c_i, c_j) \approx S_t(\bz_i, \bz_j) := \iint \Sigma_{t+1}\big((\bx', \bu^*(\bx')), (\bx'', \bu^*(\bx''))\big)\, P_t(\mathrm{d}\bx' \mid \bz_i)\, P_t(\mathrm{d}\bx'' \mid \bz_j),
    \]
    a positive semi-definite kernel on the inputs. By $|\Sigma_{t+1}(\bz', \bz'')| \le \sigma_{t+1}(\bz')\, \sigma_{t+1}(\bz'')$ and Jensen's inequality, $S_t(\bz_i, \bz_i) \le \mathbb{E}\big[\sigma^2_{t+1}(\bx', \bu^*(\bx')) \mid \bz_i\big]$, with equality when the next state is a deterministic function of $\bz_i$. Hence
    \[
        \mathrm{Cov}(\boldsymbol\eta_t) \approx D_t + S_t, \qquad D_t := \operatorname{diag}\big(\mathrm{Var}(r_i \mid \bz_i) + \mathrm{Var}(V_{t+1}(\bx') \mid \bz_i)\big)_i,
    \]
    with $S_t := \big(S_t(\bz_i, \bz_j)\big)_{ij}$. The entry $S_t(\bz_i, \bz_j)$ is close to $S_t(\bz_i, \bz_i)$ when the next states of $\bz_i$ and $\bz_j$ are close compared with the lengthscales of $\Sigma_{t+1}$, so targets at nearby inputs share most of this error.

    \paragraph{Consequence for the variances.} Under this model the posterior mean is
    \[
    \bkappa^\top (K + S_t + D_t)^{-1} \by,
    \]
    in which $S_t$ enters as a second kernel on the data, a smooth error that the regression cannot separate from $Q^*_t$ by its targets alone. Replacing $S_t$ by its diagonal, as a variance $\sigma^2_{t+1}(\bx'_i, \bu^*_i)$ added to each target's noise, would treat this shared error as independent and averaged away, which it is not. Proposition~\ref{prop:batch} therefore applies to $D_t$ alone. HGPQ sets $\Lambda_t$ to an estimate of $D_t$, namely $\sigma_r^2 + \omega_t(\bz_i)$ (Section~\ref{sec:noise}), which gives~\eqref{eq:lambda}, and leaves the shared error as a bias of the targets, which vanishes as $\mu_{t+1} \to Q^*_{t+1}$. Where $V_{t+1}$ has kinks, at states where the maximiser of $\mu_{t+1}$ switches, the smooth expansion~\eqref{eq:c_first_order} does not hold, part of the error of the next estimate behaves as independent error, and the leave-one-out estimate includes it in $\omega_t$. Appendix~\ref{app:ablation} compares this choice with adding $\sigma^2_{t+1}(\bx'_i, \bu^*_i)$ to it, with a tuned constant plus $\sigma^2_{t+1}$, and with the full covariance $S_t$ evaluated at the realised next states.

    At $t = T-1$ the next value is the terminal value $V_T$ of Section~\ref{sec:targets}, with $V^*_T = \bar r_T$ and no maximisation. Then $b_i$ is the spread of $V_T(\bx')$ over next states, which $\omega_{T-1}(\bz_i)$ absorbs together with any reward noise beyond $\sigma_r^2$. The term $c_i$ vanishes when $\bar r_T$ is known, and also when the measured terminal reward is used, since $\mathbb{E}[r_T \mid \bz_i] = \mathbb{E}[\bar r_T(\bx') \mid \bz_i]$, and the measurement noise then enters $b_i$. With a terminal GP, $c_i$ is the shared error of that GP and is treated as above. Without a terminal reward, $b_i = c_i = 0$.

\subsection{Convergence in function space}\label{app:convergence}

This subsection proves Theorem~\ref{thm:main} in a more general form, in which each target has its own variance and all variances may grow with the number of episodes by a deterministic schedule. With one variance per time step and no schedule, each backward step is the regression step of kernel value iteration \citep{yang2020provably}.

The argument has three steps. The first bounds the error of one regression step by the posterior standard deviation times a width, uniformly over the candidate value functions from which the targets are built. The second shows that coverage makes the posterior standard deviation shrink at a known rate on each region. The third adds the errors of the time steps, backwards from the last. The paragraphs below fix the notation, Appendix~\ref{app:assumptions} states the assumptions, and the lemmas of Appendix~\ref{app:proof} carry out the three steps.

\paragraph{Standing assumptions.} The set $\mathcal{X}$ is compact, $\mathcal{U}$ is a closed box with sides of positive length, $\|\cdot\|$ is the Euclidean norm on $\mathcal{Z}=\mathcal{X}\times\mathcal{U}\subset\mathbb{R}^d$ with $d:=d_x+d_u$, and the kernel $\kappa$ is continuous on $\mathcal{Z}\times\mathcal{Z}$. Hence $\kappa_{\max}:=\sup_{\mathbf{z}\in\mathcal{Z}}\kappa(\mathbf{z},\mathbf{z})<\infty$, the reproducing kernel Hilbert space (RKHS) $\mathcal{H}_\kappa$ is separable and consists of continuous functions, and its balls are relatively compact in the supremum norm on $\mathcal{Z}$. The reproducing property is $g(\mathbf{z})=\langle g,\kappa(\cdot,\mathbf{z})\rangle_\kappa$.

\paragraph{Episodes and information.} The proof tracks what the learner knows when it chooses each input. In episode $i=1,2,\dots$, time step $t$ produces a pair $\mathbf{z}_{t,i}=(\mathbf{x}_{t,i},\mathbf{u}_{t,i})$, a reward $r_{t,i}$ and a next state $\mathbf{x}_{t+1,i}$. Let $\mathcal{F}_i$ be the $\sigma$-algebra generated by everything that the learner has observed or sampled by the end of episode $i$, including its exploration noise, and let $\mathcal{G}_{t,i}$ be generated by $\mathcal{F}_{i-1}$ and everything in episode $i$ up to and including the choice of $\mathbf{u}_{t,i}$. Then $\mathbf{z}_{t,i}$ is $\mathcal{G}_{t,i}$-measurable and $(r_{t,i},\mathbf{x}_{t+1,i})$ is $\mathcal{F}_i$-measurable, hence $\mathcal{G}_{t,i+1}$-measurable.

\paragraph{Simplifications.} As in Section~\ref{sec:analysis}, rewards lie in $[0,1]$, so values at time step $t$ lie in $[0,T-t]$, the maximisation over inputs is exact, the kernel is fixed, the targets are not standardised and $N_0=0$, so the $i$-th transition of $\mathcal{D}_t$ is that of episode $i$. A known or measured terminal reward is added to the reward of the last time step and rescaled with it, $Q^*_T:=0$, and a terminal GP is not treated.

\paragraph{Bellman operator.} Let $\Pi_{[a,b]}$ denote clipping onto $[a,b]$ and $(P_tV)(\mathbf{z}):=\mathbb{E}_{\mathbf{x}'\sim P_t(\cdot\mid\mathbf{z})}[V(\mathbf{x}')]$ for bounded measurable $V$. For continuous $g:\mathcal{Z}\to\mathbb{R}$, the clipped greedy value, the Bellman operator and the targets are
\begin{equation}\label{eq:app_bellman}
\begin{gathered}
V^{t+1}_g(\mathbf{x}):=\Pi_{[0,\,T-t-1]}\Bigl(\max_{\mathbf{u}\in\mathcal{U}}g(\mathbf{x},\mathbf{u})\Bigr),\qquad
\mathcal{T}_tg:=\bar r_t+P_tV^{t+1}_g,\\
y_{t,i}(g):=r_{t,i}+V^{t+1}_g(\mathbf{x}_{t+1,i}).
\end{gathered}
\end{equation}
Clipping and maximisation are 1-Lipschitz, so for continuous $g$ and $g'$ and every $\mathbf{x}\in\mathcal{X}$,
\begin{equation}\label{eq:lip}
\bigl|V^{t+1}_g(\mathbf{x})-V^{t+1}_{g'}(\mathbf{x})\bigr|\le\sup_{\mathbf{u}\in\mathcal{U}}\bigl|g(\mathbf{x},\mathbf{u})-g'(\mathbf{x},\mathbf{u})\bigr| .
\end{equation}

\paragraph{Variances and schedule.} The target of transition $i$ at time step $t$ has a base variance $\lambda_{t,i}>0$. For a schedule exponent $\zeta\in[0,1)$ and $\chi_n:=n^{\zeta}$, the GP of time step $t$ after $n$ episodes gives this target the variance $\chi_n\lambda_{t,i}$. We write
\[
\Lambda_{t,n}:=\chi_n\operatorname{diag}(\lambda_{t,1},\dots,\lambda_{t,n}),\qquad\bar n:=n/\chi_n=n^{1-\zeta},
\]
and call $\bar n$ the effective number of episodes. The base variances in force after $n$ episodes may depend on $n$, which is not written. Algorithm~\ref{alg:hgpq} uses $\zeta=0$.

\paragraph{Posterior.} After $n$ episodes, let $K_{t,n}:=[\kappa(\mathbf{z}_{t,i},\mathbf{z}_{t,j})]_{i,j\le n}$, $\boldsymbol{\kappa}_{t,n}(\mathbf{z}):=(\kappa(\mathbf{z}_{t,i},\mathbf{z}))_{i\le n}$ and $\mathbf{y}_t(g):=(y_{t,i}(g))_{i\le n}$. For targets $\mathbf{y}\in\mathbb{R}^n$, the posterior mean and variance at time step $t$ are
\begin{equation}\label{eq:post}
\begin{aligned}
m_{t,n}[\mathbf{y}](\mathbf{z})&:=\boldsymbol{\kappa}_{t,n}(\mathbf{z})^\top(K_{t,n}+\Lambda_{t,n})^{-1}\mathbf{y},\\
\sigma^2_{t,n}(\mathbf{z})&:=\kappa(\mathbf{z},\mathbf{z})-\boldsymbol{\kappa}_{t,n}(\mathbf{z})^\top(K_{t,n}+\Lambda_{t,n})^{-1}\boldsymbol{\kappa}_{t,n}(\mathbf{z}).
\end{aligned}
\end{equation}
With exact maximisation, Algorithm~\ref{alg:hgpq} sets $\mu^n_T\equiv0$ after episode $n$ and computes, for $t=T-1,\dots,0$,
\begin{equation}\label{eq:recursion}
\mu^n_t:=m_{t,n}\bigl[\mathbf{y}_t(\mu^n_{t+1})\bigr],
\end{equation}
a finite combination of kernel functions and hence continuous. At $t=T-1$ the greedy value is clipped onto $[0,0]$, so the target is the reward alone. The error is
\[
e^n_t(\mathbf{z}):=\bigl|\mu^n_t(\mathbf{z})-Q^*_t(\mathbf{z})\bigr|,\qquad e^n_T\equiv0 .
\]

\subsubsection{Assumptions}\label{app:assumptions}

\paragraph{Sub-Gaussian noise and regions.} The noise of the targets is assumed sub-Gaussian given the past. A random variable $\eta$ is conditionally $s$-sub-Gaussian given a $\sigma$-algebra $\mathcal{G}$, for a $\mathcal{G}$-measurable $s\ge0$, if $\mathbb{E}[\exp(a\eta)\mid\mathcal{G}]\le\exp(a^2s^2/2)$ almost surely for every real $a$. Then $\mathbb{E}[\eta\mid\mathcal{G}]=0$, and $\eta/b$ is conditionally $(s/b)$-sub-Gaussian for every $\mathcal{G}$-measurable $b>0$. The error is bounded uniformly on regions $\mathcal{Z}_t:=\mathcal{X}_t\times\mathcal{U}$, with closed boxes $\mathcal{X}_t\subseteq\mathcal{X}$ whose sides have positive length.

\begin{definition}[Local approximation property]\label{def:local}
The kernel $\kappa$ has the local approximation property on a box $\mathcal{Z}_t$ with exponent $q>0$ if there are constants $L_\kappa\ge0$, $h_\kappa>0$ and $c_\kappa\ge1$ with the following property. For every $h\le h_\kappa$, every finite set $S\subset\mathcal{Z}_t$ with $\sup_{\mathbf{z}\in\mathcal{Z}_t}\min_{\mathbf{z}'\in S}\|\mathbf{z}-\mathbf{z}'\|\le h$, and every $\mathbf{z}\in\mathcal{Z}_t$, there are weights $(w_{\mathbf{z}'})_{\mathbf{z}'\in S}$ with $\sum_{\mathbf{z}'}|w_{\mathbf{z}'}|\le c_\kappa$ such that
\[
\Bigl\|\kappa(\cdot,\mathbf{z})-\sum_{\mathbf{z}'\in S}w_{\mathbf{z}'}\kappa(\cdot,\mathbf{z}')\Bigr\|_\kappa^2\le L_\kappa h^{2q}.
\]
\end{definition}

If $\kappa(\mathbf{z},\mathbf{z})+\kappa(\mathbf{z}',\mathbf{z}')-2\kappa(\mathbf{z},\mathbf{z}')\le L_\kappa\|\mathbf{z}-\mathbf{z}'\|^{2q}$ on $\mathcal{Z}_t$, the property holds with exponent $q$ and $c_\kappa=1$, by placing all of the weight on a nearest point of $S$. This H\"older condition holds with $q=1$ on bounded sets for a Mat\'ern kernel with $\nu>1$ and for the squared-exponential kernel.

\begin{assumption}\label{ass:main}
There are constants $B_t<\infty$, $\rho_t>0$, $0<\lambda_-\le\lambda_+$ and $\varsigma_t>0$, none of which depends on the episode, such that the following conditions hold for every time step $t$, every $n$ and every episode $i\le n$.
\begin{enumerate}
\item[(i)] Markov property. There is a probability kernel $\mathcal{P}_t$ from $\mathcal{Z}$ to $[0,1]\times\mathcal{X}$, under which the reward has mean $\bar r_t(\mathbf{z})$ and the next state has distribution $P_t(\cdot\mid\mathbf{z})$, such that $\mathbb{P}\bigl((r_{t,i},\mathbf{x}_{t+1,i})\in A\mid\mathcal{G}_{t,i}\bigr)=\mathcal{P}_t(A\mid\mathbf{z}_{t,i})$ almost surely for every Borel set $A$.
\item[(ii)] Completeness. For every measurable $V:\mathcal{X}\to[0,T-t-1]$, the backup $\bar r_t+P_tV$ lies in $\mathcal{H}_\kappa$ and has RKHS norm at most $B_t$.
\item[(iii)] Kernel. On every $\mathcal{Z}_t$, the kernel $\kappa$ has the local approximation property of Definition~\ref{def:local}, with an exponent $q>0$ and constants $L_\kappa$, $h_\kappa$ and $c_\kappa$ that do not depend on $t$.
\item[(iv)] Coverage and closure. For every Borel set $E\subseteq\mathcal{Z}_t$, with volume $\operatorname{Leb}(E)$, $\mathbb{P}(\mathbf{z}_{t,i}\in E\mid\mathcal{F}_{i-1})\ge\rho_t\operatorname{Leb}(E)$ almost surely. If $t\le T-2$, then $P_t(\mathcal{X}_{t+1}\mid\mathbf{z})=1$ for every $\mathbf{z}\in\mathcal{Z}_t$.
\item[(v)] Variances. The base variance $\lambda_{t,i}$ in force after $n$ episodes lies in $[\lambda_-,\lambda_+]$ and is $\mathcal{G}_{t,i}$-measurable. For every continuous $g$, the noise
$\eta_{t,i}(g):=y_{t,i}(g)-(\mathcal{T}_tg)(\mathbf{z}_{t,i})$
is conditionally $\varsigma_t\sqrt{\lambda_{t,i}}$-sub-Gaussian given $\mathcal{G}_{t,i}$.
\end{enumerate}
\end{assumption}

Condition (i) also requires that the learner's own randomness, such as its Thompson draws, carries no information about the next outcome. Condition (ii) concerns every bounded value function, not only $V^*_{t+1}$, because the targets are built from the data-dependent $\mu^n_{t+1}$. It is the completeness condition of kernel value iteration \citep{yang2020provably}, the linear MDP condition of \citet{jin2020provably} for a linear kernel, and it cannot be weakened to $Q^*_t\in\mathcal{H}_\kappa$ without losing polynomial sample complexity \citep{weisz2021exponential}. Coverage in (iv) holds with $\rho_t=\epsilon\,\upsilon_t/\Leb(\mathcal{U})$ if, given the past, the state has a density of at least $\upsilon_t>0$ on $\mathcal{X}_t$ and the uniform input is chosen with a probability $\epsilon$ that does not decay with $i$. Closure holds when the dynamics keep the states in a box $\mathcal{X}=\mathcal{X}_t$. Where the visited pairs lie on a lower-dimensional set, as for the pendulum's $(\sin\theta, \cos\theta, \dot\theta)$ or the reactor's initial $C_B = C_C = 0$, condition (iv) holds in coordinates of that set, with $d$ its dimension. Condition (v) admits a variance fixed in advance or computed from the episodes before $i$ and evaluated at $\mathbf{z}_{t,i}$, and excludes one re-estimated from the outcome of its own transition or from later episodes. Its sub-Gaussian part always holds with $\varsigma_t=(T-t)/(2\sqrt{\lambda_-})$ by Hoeffding's lemma \citep{hoeffding1963probability}, since by (i) the noise has conditional mean zero and lies in an interval of length $T-t$ with $\mathcal{G}_{t,i}$-measurable endpoints.

\paragraph{Optimal Q-functions.} If $Q^*_{t+1}$ is continuous, which holds at $t+1=T$, then $V^*_{t+1}=\max_{\mathbf{u}}Q^*_{t+1}(\cdot,\mathbf{u})$ is continuous with values in $[0,T-t-1]$, so $V^*_{t+1}=V^{t+1}_{Q^*_{t+1}}$ and the Bellman equation of Section~\ref{sec:prelim} reads
\begin{equation}\label{eq:bellmanopt}
Q^*_t=\mathcal{T}_tQ^*_{t+1} .
\end{equation}
By (ii), $Q^*_t\in\mathcal{H}_\kappa$ with norm at most $B_t$, so it is continuous, and by backward induction \eqref{eq:bellmanopt} holds for every $t\le T-1$.

\subsubsection{Main result}\label{app:main}

The rate, the burn-in and the variance constant depend on the kernel through Assumption~\ref{ass:main}(iii) and on coverage through $\rho_t$.

\begin{definition}[Rate, burn-in and variance constant]\label{def:rate}
Let $q$, $L_\kappa$, $h_\kappa$ and $c_\kappa$ be the constants of Assumption~\ref{ass:main}(iii). If $L_\kappa>0$, let $\theta:=q/(2q+d)$ and $h_n:=\bar n^{-1/(2q+d)}$. If $L_\kappa=0$, let $\theta:=1/2$ and $h_n:=h_\kappa/\sqrt{d}$, where $h_\kappa$ is taken to be at most $\sqrt{d}$ times the shortest side of every $\mathcal{Z}_t$. Let $a_t$ denote the longest side of $\mathcal{Z}_t$ and $M_{t,n}:=(2a_t/h_n)^d$. For $\delta\in(0,1)$, the burn-in $n_0(\delta)$ is the smallest integer such that, for all $n\ge n_0(\delta)$ and all $t$, $\sqrt{d}\,h_n\le h_\kappa$, every side of $\mathcal{Z}_t$ is at least $h_n$, and
\begin{equation}\label{eq:burnin}
\rho_t\,n\,h_n^d\ge2^{d+3}\log\bigl(2TM_{t,n}n^2/\delta\bigr).
\end{equation}
For $\lambda>0$ let
\[
C_{\sigma,t}(\lambda):=\begin{cases}
\bigl(L_\kappa d^q+2^{d+1}c_\kappa^2\lambda/\rho_t\bigr)^{1/2}&\text{if }L_\kappa>0,\\[2pt]
\bigl(2^{d+1}d^{d/2}c_\kappa^2\lambda/(\rho_th_\kappa^d)\bigr)^{1/2}&\text{if }L_\kappa=0,
\end{cases}
\qquad C_{\sigma,t}:=C_{\sigma,t}(\lambda_+).
\]
\end{definition}

The burn-in is finite, since the first two conditions hold for all large $n$ and the left-hand side of \eqref{eq:burnin} grows polynomially in $n$ ($n\,h_n^d\ge n^{2q/(2q+d)}$ if $L_\kappa>0$, and $n\,h_n^d\propto n$ if $L_\kappa=0$) whereas the right-hand side grows logarithmically. It grows at least exponentially with $d$.

\begin{definition}[Confidence width]\label{def:width}
For points $\mathbf{z}_1,\dots,\mathbf{z}_n\in\mathcal{Z}$ write $K_n:=[\kappa(\mathbf{z}_i,\mathbf{z}_j)]_{i,j\le n}$, and for $\lambda>0$ let
\[
\gamma_n(\lambda):=\sup_{(\mathbf{z}_1,\dots,\mathbf{z}_n)\in\mathcal{Z}^n}\tfrac12\log\det\bigl(I_n+\lambda^{-1}K_n\bigr)
\]
be the maximum information gain \citep{srinivas2010gaussian}. Let $(\tau_n)_{n\ge1}$ be a deterministic sequence of positive resolutions. For $R\ge0$, let $\mathcal{N}_n(R)$ be the logarithm of the smallest number of balls of radius $\tau_n$ in the supremum norm on $\mathcal{Z}$, with centres in $\{g\in\mathcal{H}_\kappa:\|g\|_\kappa\le R\}$, that cover this set. For $\delta\in(0,1)$, $R\ge0$ and an integer $A\ge1$, let
\begin{equation}\label{eq:width}
\bar\beta_{t,n}(\delta;R,A):=\underbrace{B_t}_{\text{function norm}}+\underbrace{2\tau_n\sqrt{\bar n/\lambda_-}}_{\text{discretisation}}+\underbrace{\varsigma_t\sqrt{2/\chi_n}\sqrt{\gamma_n(\chi_n\lambda_-)+\mathcal{N}_n(R)+\log(2Tn^2A/\delta)}}_{\text{noise}}.
\end{equation}
The confidence widths are $\beta_{T,n}(\delta):=0$ and $\beta_{t,n}(\delta):=\bar\beta_{t,n}\bigl(\delta;\beta_{t+1,n}(\delta),1\bigr)$ for $t=T-1,\dots,0$.
\end{definition}

The widths are deterministic and finite. The radius $R$ is that of a ball of candidate value functions, and $A$ is the number of variance assignments for which the bounds hold simultaneously. Since $\mathcal{N}_n(0)=0$, the width of time step $T-1$ has no covering term.

\begin{theorem}[Uniform convergence, extended form of Theorem~\ref{thm:main}]\label{thm:ext}
Let Assumption~\ref{ass:main} hold, let $\zeta\in[0,1)$ and let $\delta\in(0,1)$. With probability at least $1-2\delta$, the following hold for all time steps $t$.
\begin{enumerate}
\item[(a)] For all $n\ge1$ and all $\mathbf{z}\in\mathcal{Z}$, $\bigl|\mu^n_t(\mathbf{z})-(\mathcal{T}_t\mu^n_{t+1})(\mathbf{z})\bigr|\le\beta_{t,n}(\delta)\,\sigma_{t,n}(\mathbf{z})$.
\item[(b)] For all $n\ge1$ and all $\mathbf{z}\in\mathcal{Z}$, with the expectation taken over the next state alone,
\[
e^n_t(\mathbf{z})\le\beta_{t,n}(\delta)\,\sigma_{t,n}(\mathbf{z})+\mathbb{E}_{\mathbf{x}'\sim P_t(\cdot\mid\mathbf{z})}\Bigl[\sup_{\mathbf{u}\in\mathcal{U}}e^n_{t+1}(\mathbf{x}',\mathbf{u})\Bigr].
\]
\item[(c)] For all $n\ge n_0(\delta)$, $\displaystyle\sup_{\mathbf{z}\in\mathcal{Z}_t}e^n_t(\mathbf{z})\le\bar n^{-\theta}\sum_{j=t}^{T-1}C_{\sigma,j}\,\beta_{j,n}(\delta)$.
\end{enumerate}
If $\bar n^{-\theta}\beta_{j,n}(\delta)\to0$ as $n\to\infty$ for every $j$ and every $\delta\in(0,1)$, then $\mu^n_t\to Q^*_t$ uniformly on $\mathcal{Z}_t$ almost surely, for every $t$.
\end{theorem}

Part (b) shows that the Bellman backup passes errors on without amplification. Part (c) and the almost-sure statement form Theorem~\ref{thm:main}.

\subsubsection{Proof}\label{app:proof}

The proof bounds the error of a GP mean fitted to noisy targets by the posterior standard deviation times a width \citep{kanagawa2018gaussian,abbasi2011improved}, makes this bound uniform over a ball of candidate value functions, because the targets at time step $t$ are built from the data-dependent $\mu^n_{t+1}$, and shows that coverage makes the posterior standard deviation decrease uniformly at the rate $\bar n^{-\theta}$, a uniform version of the local variance bounds of \citet{lederer2021uniform}. Compared with kernel value iteration \citep{yang2020provably}, it replaces optimism by coverage and bounds the error of the estimate itself in the supremum norm.

\paragraph{The posterior norm.} Fix points $\mathbf{z}_1,\dots,\mathbf{z}_n\in\mathcal{Z}$ and variances $\lambda_1,\dots,\lambda_n>0$, let $\Lambda:=\operatorname{diag}(\lambda_1,\dots,\lambda_n)$, write $K$, $\boldsymbol{\kappa}_n$, $m$ and $\sigma$ for the quantities of \eqref{eq:post} built from them, and write $f(\mathbf{z}_{1:n}):=(f(\mathbf{z}_i))_{i\le n}$. The posterior norm is
\[
\|g\|_\Lambda^2:=\|g\|_\kappa^2+\sum_{i\le n}\frac{g(\mathbf{z}_i)^2}{\lambda_i},\qquad g\in\mathcal{H}_\kappa .
\]

\begin{lemma}[Posterior norm]\label{lem:norm}
For all $f,g\in\mathcal{H}_\kappa$, $\mathbf{v}\in\mathbb{R}^n$ and $\mathbf{z}\in\mathcal{Z}$,
\begin{enumerate}
\item[(a)] $|g(\mathbf{z})|\le\sigma(\mathbf{z})\,\|g\|_\Lambda$,
\item[(b)] $\|m[\mathbf{v}]\|_\Lambda^2=\mathbf{v}^\top\bigl(\Lambda^{-1}-(K+\Lambda)^{-1}\bigr)\mathbf{v}\le\|\Lambda^{-1/2}\mathbf{v}\|_2^2$,
\item[(c)] $\|f-m[f(\mathbf{z}_{1:n})]\|_\Lambda\le\|f\|_\kappa$,
\item[(d)] $\|m[f(\mathbf{z}_{1:n})]\|_\kappa\le\|f\|_\kappa$.
\end{enumerate}
Consequently, if $\mathbf{y}=f(\mathbf{z}_{1:n})+\boldsymbol{\eta}+\mathbf{b}$ with $\boldsymbol{\eta},\mathbf{b}\in\mathbb{R}^n$, then
\begin{align}
\bigl|m[\mathbf{y}](\mathbf{z})-f(\mathbf{z})\bigr|&\le\sigma(\mathbf{z})\Bigl(\|f\|_\kappa+\|m[\boldsymbol{\eta}]\|_\Lambda+\|\Lambda^{-1/2}\mathbf{b}\|_2\Bigr),\label{eq:err}\\
\|m[\mathbf{y}]\|_\kappa&\le\|f\|_\kappa+\|m[\boldsymbol{\eta}]\|_\Lambda+\|\Lambda^{-1/2}\mathbf{b}\|_2.\label{eq:nrm}
\end{align}
\end{lemma}
Lemma~\ref{lem:norm} reduces the error of a posterior mean at any pair to the posterior standard deviation times three norms, which become the function-norm, noise and discretisation terms of the width $\beta_{t,n}(\delta)$ in Theorem~\ref{thm:ext}(a).

\begin{proof}
Part (a). For weights $\mathbf{w}\in\mathbb{R}^n$, the reproducing property gives $g(\mathbf{z})=\langle g,\kappa(\cdot,\mathbf{z})-\sum_iw_i\kappa(\cdot,\mathbf{z}_i)\rangle_\kappa+\sum_i(g(\mathbf{z}_i)/\sqrt{\lambda_i})\sqrt{\lambda_i}\,w_i$, and the Cauchy--Schwarz inequality bounds $|g(\mathbf{z})|^2$ by $\|g\|_\Lambda^2$ times
\begin{equation}\label{eq:phi}
\Phi_{\mathbf{z}}(\mathbf{w}):=\Bigl\|\kappa(\cdot,\mathbf{z})-\sum_iw_i\kappa(\cdot,\mathbf{z}_i)\Bigr\|_\kappa^2+\sum_i\lambda_iw_i^2=\kappa(\mathbf{z},\mathbf{z})-2\mathbf{w}^\top\boldsymbol{\kappa}_n(\mathbf{z})+\mathbf{w}^\top(K+\Lambda)\mathbf{w}.
\end{equation}
Its minimum, at $\mathbf{w}=(K+\Lambda)^{-1}\boldsymbol{\kappa}_n(\mathbf{z})$, is $\sigma^2(\mathbf{z})$. Hence
\begin{equation}\label{eq:sigmaphi}
|g(\mathbf{z})|^2\le\sigma^2(\mathbf{z})\,\|g\|_\Lambda^2\qquad\text{and}\qquad\sigma^2(\mathbf{z})\le\Phi_{\mathbf{z}}(\mathbf{w})\ \text{ for every }\mathbf{w}\in\mathbb{R}^n .
\end{equation}

Parts (b) to (d) are standard facts of kernel ridge regression \citep{kanagawa2018gaussian}.

Part (b). The mean $m[\mathbf{v}]=\sum_ic_i\kappa(\cdot,\mathbf{z}_i)$ is a combination of kernel functions with weights $\mathbf{c}:=(K+\Lambda)^{-1}\mathbf{v}$, so its posterior norm is
\[
\|m[\mathbf{v}]\|_\Lambda^2=\mathbf{c}^\top K\mathbf{c}+\mathbf{c}^\top K\Lambda^{-1}K\mathbf{c}=\mathbf{v}^\top(K+\Lambda)^{-1}K\Lambda^{-1}\mathbf{v},
\]
which equals the stated form. The inequality holds because $(K+\Lambda)^{-1}\succeq0$.

Parts (c) and (d). Let $m_f:=m[f(\mathbf{z}_{1:n})]$ be the posterior mean of $f$ observed without noise, $\tilde f:=f-m_f$ its residual and $\mathbf{c}:=(K+\Lambda)^{-1}f(\mathbf{z}_{1:n})$ its weights. The residual at the data is $\tilde f(\mathbf{z}_{1:n})=\Lambda\mathbf{c}$, so
\begin{equation}\label{eq:fitinner}
\langle m_f,g\rangle_\kappa=\sum_ic_ig(\mathbf{z}_i)=\sum_i\frac{\tilde f(\mathbf{z}_i)\,g(\mathbf{z}_i)}{\lambda_i}\qquad\text{for every }g\in\mathcal{H}_\kappa.
\end{equation}
With $g=\tilde f$, the right-hand side is the data term of $\|\tilde f\|_\Lambda^2$, which gives (c),
\[
\|\tilde f\|_\Lambda^2=\|\tilde f\|_\kappa^2+\langle m_f,\tilde f\rangle_\kappa=\langle f,\tilde f\rangle_\kappa\le\|f\|_\kappa\|\tilde f\|_\Lambda .
\]
The same choice shows $\langle m_f,\tilde f\rangle_\kappa\ge0$, which gives (d),
\[
\|m_f\|_\kappa^2=\langle m_f,f\rangle_\kappa-\langle m_f,\tilde f\rangle_\kappa\le\langle m_f,f\rangle_\kappa\le\|m_f\|_\kappa\|f\|_\kappa .
\]

Consequences. By linearity, $m[\mathbf{y}]-f=m[\boldsymbol{\eta}]+m[\mathbf{b}]-\tilde f$. Part (a) applied to each term, with $\|m[\mathbf{b}]\|_\Lambda\le\|\Lambda^{-1/2}\mathbf{b}\|_2$ from (b) and $\|\tilde f\|_\Lambda\le\|f\|_\kappa$ from (c), proves \eqref{eq:err}. Similarly, \eqref{eq:nrm} follows from $m[\mathbf{y}]=m_f+m[\boldsymbol{\eta}]+m[\mathbf{b}]$, part (d), $\|\cdot\|_\kappa\le\|\cdot\|_\Lambda$ and part (b).
\end{proof}

\begin{lemma}[Noise term]\label{lem:noise}
Let $(\mathcal{G}_i)_{i\ge1}$ be a filtration and let $\lambda_{\min}>0$ and $\varsigma>0$ be constants. For $i\le n$, let $\mathbf{z}_i$ and $\lambda_i\ge\lambda_{\min}$ be $\mathcal{G}_i$-measurable, and let $\eta_i$ be $\mathcal{G}_{i+1}$-measurable and conditionally $\varsigma\sqrt{\lambda_i}$-sub-Gaussian given $\mathcal{G}_i$. With probability at least $1-\delta$,
\[
\|m[\eta_{1:n}]\|_\Lambda\le\varsigma\sqrt{2\gamma_n(\lambda_{\min})+2\log(1/\delta)} .
\]
\end{lemma}
Lemma~\ref{lem:noise} supplies the noise term of the width $\beta_{t,n}(\delta)$ in Theorem~\ref{thm:ext}(a).

\begin{proof}
The proof rescales each noise term by its standard deviation and applies a self-normalised bound. Let $\tilde\eta_i:=\eta_i/\sqrt{\lambda_i}$, which is conditionally $\varsigma$-sub-Gaussian given $\mathcal{G}_i$ because $\lambda_i$ is $\mathcal{G}_i$-measurable, and let $\tilde K:=\Lambda^{-1/2}K\Lambda^{-1/2}$. With the features $\phi_i:=\kappa(\cdot,\mathbf{z}_i)/\sqrt{\lambda_i}$, which are $\mathcal{G}_i$-measurable, and $F\mathbf{a}:=\sum_ia_i\phi_i$, so that $F^*F=\tilde K$, Lemma~\ref{lem:norm}(b) gives
\[
\|m[\boldsymbol{\eta}]\|_\Lambda^2=\tilde{\boldsymbol{\eta}}^\top\tilde K(\tilde K+I)^{-1}\tilde{\boldsymbol{\eta}}=\bigl\langle F\tilde{\boldsymbol{\eta}},(FF^*+\mathrm{id})^{-1}F\tilde{\boldsymbol{\eta}}\bigr\rangle_\kappa .
\]
The self-normalised inequality of \citet{abbasi2011improved} in the separable Hilbert space $\mathcal{H}_\kappa$ \citep[Theorem~1, with regulariser one]{whitehouse2023sublinear} bounds it by $2\varsigma^2\log(\det(I+\tilde K)^{1/2}/\delta)$ with probability at least $1-\delta$. Finally, $\det(I+\tilde K)=\det(I+K^{1/2}\Lambda^{-1}K^{1/2})\le\det(I+\lambda_{\min}^{-1}K)\le e^{2\gamma_n(\lambda_{\min})}$, since $\Lambda^{-1}\preceq\lambda_{\min}^{-1}I$.
\end{proof}

\paragraph{Uniform concentration.} A variance assignment is a family of base variances $(\lambda_{t,i})$, for all $t$ and $i\le n$, that satisfies Assumption~\ref{ass:main}(v). In Theorem~\ref{thm:ext} there is one, that of the algorithm.

\begin{lemma}[Uniform concentration]\label{lem:conc}
Let Assumption~\ref{ass:main}(i) and (ii) hold. For every $n$, let $R_{1,n},\dots,R_{T,n}\ge0$ be deterministic radii, let $A_n\ge1$ be a deterministic integer, and let $A_n$ variance assignments be given, indexed by $j\le A_n$. Write $m^{(j)}_{t,n}$ and $\sigma^{(j)}_{t,n}$ for the posterior mean and standard deviation \eqref{eq:post} with the base variances of assignment $j$. With probability at least $1-\delta$, for all $t$, all $n\ge1$, all $j\le A_n$, all $g\in\mathcal{H}_\kappa$ with $\|g\|_\kappa\le R_{t+1,n}$ and all $\mathbf{z}\in\mathcal{Z}$,
\begin{enumerate}
\item[(a)] $\bigl|m^{(j)}_{t,n}[\mathbf{y}_t(g)](\mathbf{z})-(\mathcal{T}_tg)(\mathbf{z})\bigr|\le\bar\beta_{t,n}(\delta;R_{t+1,n},A_n)\,\sigma^{(j)}_{t,n}(\mathbf{z})$,
\item[(b)] $\bigl\|m^{(j)}_{t,n}[\mathbf{y}_t(g)]\bigr\|_\kappa\le\bar\beta_{t,n}(\delta;R_{t+1,n},A_n)$.
\end{enumerate}
\end{lemma}
Lemma~\ref{lem:conc} gives Theorem~\ref{thm:ext}(a) and the bound $\|\mu^n_t\|_\kappa \le \beta_{t,n}(\delta)$ that carries the argument from one time step to the one before it.

\begin{proof}
The proof applies Lemma~\ref{lem:noise} to every function of a finite cover of the ball and passes from the cover to the whole ball, which costs the discretisation term of the width. Fix $t$, $n$ and $j$, and let $R:=R_{t+1,n}$. Choose a deterministic set $\mathcal{C}$ of $e^{\mathcal{N}_n(R)}$ centres in the ball $\{\|g\|_\kappa\le R\}$ that covers it at resolution $\tau_n$ in the supremum norm.

For a centre $g$, $y_{t,i}(g)=(\mathcal{T}_tg)(\mathbf{z}_{t,i})+\eta_{t,i}(g)$. The variances $\chi_n\lambda_{t,i}\ge\chi_n\lambda_-$ are $\mathcal{G}_{t,i}$-measurable, and by Assumption~\ref{ass:main}(v) $\eta_{t,i}(g)$ is $\mathcal{G}_{t,i+1}$-measurable and conditionally $(\varsigma_t/\sqrt{\chi_n})\sqrt{\chi_n\lambda_{t,i}}$-sub-Gaussian given $\mathcal{G}_{t,i}$. Lemma~\ref{lem:noise} with $\varsigma=\varsigma_t/\sqrt{\chi_n}$, $\lambda_{\min}=\chi_n\lambda_-$ and confidence level $\delta/(2Tn^2A_n|\mathcal{C}|)$ gives
\begin{equation}\label{eq:noisebound}
\bigl\|m^{(j)}_{t,n}[\boldsymbol{\eta}_t(g)]\bigr\|_{\Lambda_{t,n}}\le\varsigma_t\sqrt{2/\chi_n}\,\sqrt{\gamma_n(\chi_n\lambda_-)+\mathcal{N}_n(R)+\log(2Tn^2A_n/\delta)},
\end{equation}
with $\boldsymbol{\eta}_t(g):=(\eta_{t,i}(g))_{i\le n}$. Since $\sum_nn^{-2}<2$, a union bound gives \eqref{eq:noisebound} for every centre, assignment, $n$ and $t$ with probability at least $1-\delta$.

For $g$ in the ball, let $g_c\in\mathcal{C}$ lie within $\tau_n$ of $g$. Then $\mathbf{y}_t(g)=\bigl((\mathcal{T}_tg)(\mathbf{z}_{t,i})\bigr)_{i\le n}+\boldsymbol{\eta}_t(g_c)+\mathbf{b}$ with $b_i:=\bigl(V^{t+1}_g-V^{t+1}_{g_c}\bigr)(\mathbf{x}_{t+1,i})-\bigl(P_t\bigl(V^{t+1}_g-V^{t+1}_{g_c}\bigr)\bigr)(\mathbf{z}_{t,i})$. By \eqref{eq:lip}, $|b_i|\le2\tau_n$, so $\|\Lambda_{t,n}^{-1/2}\mathbf{b}\|_2\le2\tau_n\sqrt{n/(\chi_n\lambda_-)}=2\tau_n\sqrt{\bar n/\lambda_-}$. With $\|\mathcal{T}_tg\|_\kappa\le B_t$ by Assumption~\ref{ass:main}(ii), the three terms of \eqref{eq:err} and \eqref{eq:nrm}, with $f=\mathcal{T}_tg$ and $\boldsymbol{\eta}=\boldsymbol{\eta}_t(g_c)$, are bounded by those of \eqref{eq:width}, which proves (a) and (b).
\end{proof}

\begin{lemma}[Variance decay]\label{lem:var}
Let Assumption~\ref{ass:main}(iii) and (iv) hold. There is an event of probability at least $1-\delta$, determined by the visited pairs alone, on which, for all $t$, all $n\ge n_0(\delta)$, all $\mathbf{z}\in\mathcal{Z}_t$ and every choice of base variances in $(0,\lambda_+]$, where $\sigma_{t,n}$ is computed with these base variances and the schedule, $\sigma_{t,n}(\mathbf{z})\le C_{\sigma,t}\,\bar n^{-\theta}$.
\end{lemma}
Lemma~\ref{lem:var} supplies the factor $C_{\sigma,t}\,\bar n^{-\theta}$ in Theorem~\ref{thm:ext}(c).

\begin{proof}
The proof shows that every cell of a grid on $\mathcal{Z}_t$ receives many visited pairs, and then combines kernel functions at these pairs into an approximation of $\kappa(\cdot,\mathbf{z})$ whose weights are small. Fix $t$ and $n\ge n_0(\delta)$. Splitting each side of $\mathcal{Z}_t$ into equal intervals of length between $h_n/2$ and $h_n$ partitions $\mathcal{Z}_t$ into at most $M_{t,n}$ cells of diameter at most $\sqrt{d}\,h_n$ and volume at least $(h_n/2)^d$.

For a cell, let $X_i$ be the indicator that $\mathbf{z}_{t,i}$ lies in it, which is $\mathcal{F}_i$-measurable, and let $p_n:=\rho_t(h_n/2)^d$. By coverage, $\mathbb{P}(X_i=1\mid\mathcal{F}_{i-1})\ge p_n$, so $\mathbb{E}[e^{-vX_i}\mid\mathcal{F}_{i-1}]\le\exp(-p_n(1-e^{-v}))$ for $v>0$. By induction, and by Markov's inequality with $v=\log2$,
\begin{gather*}
\mathbb{E}\Bigl[e^{-v\sum_{i\le n}X_i}\Bigr]\le\exp\bigl(-np_n(1-e^{-v})\bigr),\\
\mathbb{P}\Bigl(\sum_{i\le n}X_i\le np_n/2\Bigr)\le\exp\bigl(-np_n(1-\log2)/2\bigr)\le e^{-np_n/8}.
\end{gather*} By \eqref{eq:burnin} and union bounds over the cells, $t$ and $n$, every cell receives more than $np_n/2$ pairs for all $t$ and $n\ge n_0(\delta)$, except on an event of probability at most $\delta$ that does not involve the variances.

On this event, let $J>np_n/2$ be the smallest cell count, and choose $J$ disjoint layers of visited pairs, each containing one pair of every cell. Every point of $\mathcal{Z}_t$ lies within $\sqrt{d}\,h_n\le h_\kappa$ of each layer. Fix $\mathbf{z}\in\mathcal{Z}_t$ and base variances in $(0,\lambda_+]$, let $\mathbf{w}^{(k)}$ be the weights of Definition~\ref{def:local} for layer $k$ with $h=\sqrt{d}\,h_n$, and in \eqref{eq:phi}, with variances $\chi_n\lambda_{t,i}$, put the weight $w^{(k)}_{\mathbf{z}'}/J$ on each pair $\mathbf{z}'$ of layer $k$ and zero elsewhere. By convexity of the squared norm, the kernel part of $\Phi_{\mathbf{z}}$ is at most $L_\kappa d^qh_n^{2q}$. Since the layers are disjoint and $\sum_{\mathbf{z}'}|w^{(k)}_{\mathbf{z}'}|\le c_\kappa$, the noise part is at most $\chi_n\lambda_+\sum_{k\le J}c_\kappa^2/J^2=c_\kappa^2\chi_n\lambda_+/J$. By \eqref{eq:sigmaphi}, $J>\rho_t(h_n/2)^dn/2$ and $\bar n=n/\chi_n$,
\[
\sigma^2_{t,n}(\mathbf{z})\le L_\kappa d^qh_n^{2q}+\frac{2^{d+1}c_\kappa^2\lambda_+}{\rho_t\,h_n^d\,\bar n}.
\]
If $L_\kappa>0$, then $h_n^{2q}=1/(h_n^d\bar n)=\bar n^{-2\theta}$. If $L_\kappa=0$, the first term vanishes, $h_n^d=h_\kappa^dd^{-d/2}$ and $2\theta=1$. In both cases $\sigma^2_{t,n}(\mathbf{z})\le C_{\sigma,t}^2\,\bar n^{-2\theta}$.
\end{proof}

\begin{lemma}[Error propagation]\label{lem:prop}
Let Assumption~\ref{ass:main}(ii) hold and fix $n$. For all $t\le T-1$ and all $\mathbf{z}\in\mathcal{Z}$, with the expectation taken over the next state alone,
\begin{equation}\label{eq:prop}
e^n_t(\mathbf{z})\le\bigl|\mu^n_t(\mathbf{z})-(\mathcal{T}_t\mu^n_{t+1})(\mathbf{z})\bigr|+\mathbb{E}_{\mathbf{x}'\sim P_t(\cdot\mid\mathbf{z})}\Bigl[\sup_{\mathbf{u}\in\mathcal{U}}e^n_{t+1}(\mathbf{x}',\mathbf{u})\Bigr].
\end{equation}
If in addition the closure condition of Assumption~\ref{ass:main}(iv) holds and $\sup_{\mathbf{z}\in\mathcal{Z}_t}|\mu^n_t(\mathbf{z})-(\mathcal{T}_t\mu^n_{t+1})(\mathbf{z})|\le\iota_t$ for every $t\le T-1$, then
\[
\sup_{\mathbf{z}\in\mathcal{Z}_t}e^n_t(\mathbf{z})\le\sum_{j=t}^{T-1}\iota_j\qquad\text{for every }t\le T-1 .
\]
\end{lemma}
Lemma~\ref{lem:prop} gives Theorem~\ref{thm:ext}(b) and adds the errors of the time steps into the sum of Theorem~\ref{thm:ext}(c).

\begin{proof}
Since $e^n_{t+1}$ is continuous on the compact $\mathcal{Z}$, $\mathbf{x}'\mapsto\sup_{\mathbf{u}}e^n_{t+1}(\mathbf{x}',\mathbf{u})$ is continuous and the expectation in \eqref{eq:prop} is well defined. The error at time step $t$ splits into the error of one regression step and the error passed on from time step $t+1$. By \eqref{eq:bellmanopt} and the triangle inequality,
\[
e^n_t(\mathbf{z})\le\bigl|\mu^n_t(\mathbf{z})-(\mathcal{T}_t\mu^n_{t+1})(\mathbf{z})\bigr|+\bigl|(\mathcal{T}_t\mu^n_{t+1}-\mathcal{T}_tQ^*_{t+1})(\mathbf{z})\bigr|,
\]
and by \eqref{eq:lip} the second term is at most the expectation in \eqref{eq:prop}. At $t=T-1$ it vanishes, since both greedy values are clipped onto $[0,0]$. For the second claim, let $\Delta_t:=\sup_{\mathbf{z}\in\mathcal{Z}_t}e^n_t(\mathbf{z})$ and $\Delta_T:=0$. For $t\le T-2$ and $\mathbf{z}\in\mathcal{Z}_t$, closure keeps the next state in $\mathcal{X}_{t+1}$ almost surely, so the expectation in \eqref{eq:prop} is at most $\Delta_{t+1}$. Hence $\Delta_t\le\iota_t+\Delta_{t+1}$ for every $t\le T-1$, and summing gives the claim.
\end{proof}

\begin{proof}[Proof of Theorem~\ref{thm:ext}]
The proof combines the three steps. Apply Lemma~\ref{lem:conc} with $A_n=1$, the variance assignment of the algorithm and the radii $R_{t,n}:=\beta_{t,n}(\delta)$, so that its width is $\beta_{t,n}(\delta)$, and work on the event, of probability at least $1-2\delta$, on which it and Lemma~\ref{lem:var} hold.

Part (a). Fix $n$. By backward induction, $\|\mu^n_t\|_\kappa\le\beta_{t,n}(\delta)$ for all $t$. This holds at $t=T$. If it holds at $t+1$, Lemma~\ref{lem:conc}, being uniform over the ball of radius $\beta_{t+1,n}(\delta)$, applies to the data-dependent $g=\mu^n_{t+1}$, and by \eqref{eq:recursion} its part (b) is the norm bound at $t$ and its part (a) is the claim. Part (b) follows from (a) and \eqref{eq:prop}. For $n\ge n_0(\delta)$, part (c) follows from (a), Lemma~\ref{lem:var} and Lemma~\ref{lem:prop} with $\iota_t=C_{\sigma,t}\beta_{t,n}(\delta)\bar n^{-\theta}$.

Almost-sure convergence. Let $\mathcal{E}$ be the event that $\mu^n_t\to Q^*_t$ uniformly on $\mathcal{Z}_t$ for every $t$. Under the condition of the theorem, (c) shows that $\mathcal{E}$ contains an event of probability at least $1-2\delta$ for every $\delta\in(0,1)$, so $\mathbb{P}(\mathcal{E})=1$.
\end{proof}

\subsubsection{Rates}\label{app:consequences}

The widths $\beta_{t,n}(\delta)$ grow with $\gamma_n$ and $\mathcal{N}_n$, and the corollary below turns polynomial bounds on both into a rate. We write $f_n(\delta)=\tilde O(g_n)$ if $f_n(\delta)\le C\,g_n\log^p(en/\delta)$ for constants $C$ and $p\ge0$ that depend on neither $n$ nor $\delta$.

\begin{corollary}[Rates under polynomial complexity]\label{cor:rates}
Let Assumption~\ref{ass:main} hold, and suppose that there is an exponent $\psi\in(0,1)$ with the following two properties.
\begin{enumerate}
\item[(P1)] There are constants $C_\gamma$ and $p_\gamma\ge0$ such that $\gamma_n(\lambda)\le C_\gamma\max\{1,(n/\lambda)^{\psi}\}\log^{p_\gamma}(e+n/\lambda)$ for all $n\ge1$ and all $\lambda\ge\lambda_-$.
\item[(P2)] There is a constant $C_{\mathcal{N}}$ such that, for all $0<\tau\le R$, the ball $\{g\in\mathcal{H}_\kappa:\|g\|_\kappa\le R\}$ is covered by $\exp\bigl(C_{\mathcal{N}}(R/\tau)^{2\psi}\bigr)$ balls of radius $\tau$ in the supremum norm on $\mathcal{Z}$ with centres in it.
\end{enumerate}
Let $\zeta\in[0,1)$, let $D_n:=\max\{1,\bar n^{\psi}/\chi_n\}$ and choose the resolution $\tau_n:=\sqrt{D_n/\bar n}$. Then $\beta_{t,n}(\delta)=\tilde O(\sqrt{D_n})$ for every $t$. Moreover, there are constants $C$ and $p\ge0$ such that, for every $\delta\in(0,1)$, with probability at least $1-2\delta$, for all $n\ge n_0(\delta)$ and all $t$,
\[
\sup_{\mathbf{z}\in\mathcal{Z}_t}e^n_t(\mathbf{z})\le C\,n^{-\xi(\zeta)}\log^p(en/\delta),\qquad\xi(\zeta):=(1-\zeta)\,\theta-\tfrac12\max\bigl\{0,\,(1-\zeta)\,\psi-\zeta\bigr\}.
\]
The exponent $\xi(\zeta)$ is largest at $\zeta^*:=\psi/(1+\psi)$, where $D_n=1$ and
\[
\xi(\zeta^*)=\frac{\theta}{1+\psi}>0,\qquad\text{whereas}\qquad\xi(0)=\theta-\frac{\psi}{2}.
\]
If $\xi(\zeta)>0$, then $\mu^n_t\to Q^*_t$ uniformly on $\mathcal{Z}_t$ almost surely, for every $t$.
\end{corollary}

\begin{proof}
Each of the three terms of the width grows at most as $\sqrt{D_n}$, up to logarithmic factors. By backward induction on $t$, $\beta_{t,n}(\delta)=\tilde O(\sqrt{D_n})$. The function-norm term of \eqref{eq:width} is constant, and the discretisation term equals $2\sqrt{D_n/\lambda_-}$ by the choice of $\tau_n$. In the noise term, (P1) gives $\gamma_n(\chi_n\lambda_-)=\tilde O(\bar n^{\psi})$, and (P2) with $R=\beta_{t+1,n}(\delta)=\tilde O(\sqrt{D_n})$ gives $\mathcal{N}_n(R)=\tilde O(\bar n^{\psi})$, or $\mathcal{N}_n(R)\le C_{\mathcal{N}}$ if $R<\tau_n$. Since $\bar n^{\psi}/\chi_n\le D_n$, the noise term is $\tilde O(\sqrt{D_n})$. Theorem~\ref{thm:ext}(c) then bounds the error by $\tilde O(\sqrt{D_n}\,\bar n^{-\theta})=\tilde O(n^{-\xi(\zeta)})$, since $\bar n=n^{1-\zeta}$ and $\bar n^{\psi}/\chi_n=n^{(1-\zeta)\psi-\zeta}$. Because $\theta\le1/2$, the exponent $\xi$ increases in $\zeta$ up to $\zeta^*$, where $(1-\zeta^*)\psi=\zeta^*$ and $D_n=1$, and decreases beyond it. If $\xi(\zeta)>0$, then $\bar n^{-\theta}\beta_{j,n}(\delta)\to0$ and the almost-sure statement of Theorem~\ref{thm:ext} applies.
\end{proof}

\paragraph{Mat\'ern kernels.} Let $\kappa$ be a Mat\'ern kernel with smoothness $\nu>1/2$ and let $\psi:=d/(2\nu+d)$. Property (P2) holds with this $\psi$, because $\mathcal{H}_\kappa$ is norm-equivalent to the Sobolev space of order $\nu+d/2$ on $\mathcal{Z}$ \citep{kanagawa2018gaussian}, whose balls have metric entropy of order $(R/\tau)^{d/(\nu+d/2)}$ in the supremum norm \citep{edmunds1996function}. Property (P1) holds with the same $\psi$ if the eigenfunctions of the kernel are uniformly bounded \citep{vakili2021information}. With $L_\kappa>0$, Corollary~\ref{cor:rates} gives
\[
\xi(0)=\frac{4q\nu-d^2}{2\,(2q+d)\,(2\nu+d)}\qquad\text{and}\qquad\xi(\zeta^*)=\frac{q\,(2\nu+d)}{(2q+d)\,(2\nu+2d)}\quad\text{with}\quad\zeta^*=\frac{d}{2\nu+2d}.
\]
Without a schedule the bound tends to zero if $d^2<4q\nu$, so for $d\le3$ with the Mat\'ern-5/2 kernel ($\nu=5/2$, $q=1$). With the schedule $\chi_n=n^{d/(2\nu+2d)}$ it tends to zero in every dimension and the confidence width stays polylogarithmic. \citet{whitehouse2023sublinear} use the same device, a regulariser that grows as a power of the number of observations, to obtain sublinear regret for GP-UCB with every Mat\'ern kernel.

\section{Implementation details}\label{app:implementation}

    \paragraph{Terminal value and clipping.}\label{app:terminal} For $t < T-1$ the maximum in the target~\eqref{eq:target} is clipped to the range of returns from time step $t+1$, $V_{t+1}(\bx) := \Pi_{t+1}\big(\max_{\bu \in \mathcal{U}} \mu_{t+1}(\bx, \bu)\big)$, where $\Pi_{t+1}$ clips to the interval from $(T-t-1)\, r_{\min} + r^T_{\min}$ to $(T-t-1)\, r_{\max} + r^T_{\max}$, so that an overestimated maximum does not propagate backwards. The terminal value is
    \begin{equation}\label{eq:terminal}
    V_T(\bx'_i) := \bar r_T(\bx'_i), \qquad V_T(\bx'_i) := r_{T,i}, \qquad \text{or} \qquad V_T(\bx'_i) := \Pi_T\big(\mu_T(\bx'_i)\big),
    \end{equation}
    The first applies when $\bar r_T$ is a known function of the state, as in the synthetic MDPs (Appendix~\ref{app:toy}). The second uses the terminal reward $r_{T,i}$ measured in the episode of transition $i$, which is equivalent to adding it to $r_{T-1}$. The third fits a GP $R_T \sim \mathcal{GP}(\mu_T, \Sigma_T)$ on the $d_x$ state coordinates to past measured pairs $(\bx_T, r_T)$, with $\Pi_T$ clipping to $[r^T_{\min}, r^T_{\max}]$, which pools the measurements of nearby final states. Without a terminal reward, $V_T \equiv 0$, as in the five benchmarks, which give the reward of the final state on the last transition (Appendix~\ref{app:models}).

    \paragraph{Fitting the GPs.}\label{app:fitting} After each episode the GP of every time step is refitted. Inputs are scaled to $[-1, 1]$ by the state and input bounds, and targets are standardised to zero mean and unit variance. The Mat\'ern-5/2 kernel has a signal variance in $[0.1, 10]$ and one lengthscale per input dimension in $[0.5, 10]$, fitted by marginal likelihood. The variances $\omega_t(\bz_i)$ start from a single level fitted by marginal likelihood with the kernel and are then estimated as in the most likely heteroscedastic GP of \citet{kersting2007most}. With the current variances, the leave-one-out residual $e_i = y_{t,i} - \mu_{t,-i}(\bz_i)$ and its predictive variance $s_i^2$ follow in closed form from one inverse of $K + \Lambda_t$ \citep[Section~5.4.2]{rasmussen2006gaussian}. A second GP, of the same kernel family with its own hyperparameters and Gaussian noise level fitted by marginal likelihood, regresses $\log e_i^2$ on the inputs. With $m(\bz)$ its posterior mean, the update is $\omega_t(\bz_i) \leftarrow \exp(m(\bz_i) + 1.27) - (s_i^2 - \omega_t(\bz_i))$, clipped to $[10^{-6}, 1]$ in standardised units. Here $\exp(m + 1.27)$ estimates the expected squared residual, with $1.27 = -\mathbb{E}[\log \chi^2_1]$ the bias of the logarithm, so $m_t$ of Section~\ref{sec:noise} is $m + 1.27$, and $s_i^2 - \omega_t(\bz_i)$ is the GP's leave-one-out predictive variance plus the floor $\sigma_r^2$. The kernel is then refitted with these variances fixed, and the two steps alternate twice. For the sparse GP of Appendix~\ref{app:scaling}, the leave-one-out quantities follow by the Woodbury identity at cost $O(nm^2)$, and the GP on the log residuals is sparse as well. On the LMDP the kernel is linear in the features of Appendix~\ref{app:hparams} and fixed.

    \paragraph{Behaviour policy and maximisation.} Each of the $N_0$ initial episodes (Section~\ref{sec:setup}) applies the inputs of its own scrambled Sobol sequence over $\mathcal{U}$. Thompson samples $\tilde Q_t$ are drawn as whole functions by pathwise conditioning \citep{wilson2020efficiently}, a random-Fourier-feature prior sample \citep{rahimi2007random} corrected by Matheron's rule, so they are smooth with analytic gradients. Every maximisation over inputs (of $\tilde Q_t$ when acting, of $\mu_{t+1}$ for the targets and of $\mu_t$ for the greedy policy) evaluates the function on a fixed candidate set, runs L-BFGS-B \citep{byrd1995limited} over $\mathcal{U}$ from the best candidates, and returns the best input found (settings in Appendix~\ref{app:hparams}).

    \paragraph{Computational cost.}\label{app:scalability} After $n$ episodes a backward pass solves $T$ linear systems of size $N_0 + n$, at cost $O(T (N_0 + n)^3)$, plus the maximisations for the targets. For our budgets ($n \le 300$) this is small compared with a physical experiment. Appendix~\ref{app:scaling} measures the cost of the exact fit and of a sparse fit \citep{titsias2009variational} for larger budgets.
    
    \section{Benchmark models}\label{app:models}
    
    Table~\ref{tab:env_dims} summarises the five benchmarks.
    
    \begin{table}[h]
    \centering
    \caption{Benchmarks: state dimension $d_x$, input dimension $d_u$, horizon $T$ and reward.}
    \label{tab:env_dims}
    \small
    \renewcommand{\arraystretch}{1.1}
    \begin{tabular}{lcccl}
    \toprule
    \textbf{Benchmark} & $d_x$ & $d_u$ & $T$ & \textbf{Reward} \\
    \midrule
    Non-stationary LMDP               & 2 & 1 & 10 & every time step, linear, sign-switching \\
    Semi-batch reactor                & 5 & 2 & 8  & final time step only (product) \\
    Photosynthetic biomass production & 3 & 2 & 8  & final time step only (biomass) \\
    Inverted pendulum                 & 3 & 1 & 20 & final state, small input cost \\
    Target tracking                   & 6 & 2 & 10 & final distance, small input cost \\
    \bottomrule
    \end{tabular}
    \end{table}
    
    \subsection{Non-stationary linear MDP}
    
    The state $\bx_t = [p_t, v_t]^\top$ evolves as
    \begin{equation}
    \bx_{t+1} = A\bx_t + B\bu_t + \bw_t, \qquad
    A = \begin{bmatrix} 1 & \Delta t \\ 0 & 1 \end{bmatrix}, \quad B = \begin{bmatrix} \tfrac12 \Delta t^2 \\ \Delta t \end{bmatrix},
    \end{equation}
    with input $\bu_t \in [-1, 1]$, time step $\Delta t = 1.2$ and disturbance $\bw_t \sim \mathcal{N}(\mathbf{0}, \sigma_w^2 I)$, $\sigma_w = 0.1$. The horizon is $T = 10$. The reward is linear in the state and the input, with a random coefficient $\beta_t$ on the input,
    \begin{equation}
    r_t = -\big( A_r \bx_t + B_r\, \beta_t\, \bu_t \big), \qquad A_r = [0.02,\ 0], \quad B_r = 1,
    \end{equation}
    and the coefficient changes sign every two time steps,
    \begin{equation}
    \beta_t \sim
    \begin{cases}
    \mathcal{N}(+1, \sigma_\beta^2) & \text{if } \lfloor t/2 \rfloor \bmod 2 = 0, \\
    \mathcal{N}(-1, \sigma_\beta^2) & \text{otherwise,}
    \end{cases}
    \qquad \sigma_\beta = 1,
    \end{equation}
    so the controller must anticipate the reversals. Rewards are divided by 100 before learning. Initial states are drawn as $p_0 \sim \mathcal{U}(3, 7)$ and $v_0 \sim \mathcal{U}(-2, 2)$. 
    
    \subsection{Semi-batch reactor}
    
    A semi-batch reactor carries out two consecutive reactions,
    \begin{align}
        2A &\xrightarrow{k_1} B, \\
        B &\xrightarrow{k_2} 3C.
    \end{align}
    The state $\bx = [C_A, C_B, C_C, T_r, V]^\top$ tracks the species concentrations, the reactor temperature $T_r$ and the liquid volume $V$. The mass balances are
    \begin{align}
        \dot{C}_A &= -k_1 C_A + \frac{F}{V}(C_{A0} - C_A), \\
        \dot{C}_B &= \tfrac{1}{2} k_1 C_A - k_2 C_B - \frac{F}{V} C_B, \\
        \dot{C}_C &= 3 k_2 C_B - \frac{F}{V} C_C, \\
        \dot{V} &= F,
    \end{align}
    where $F$ is the feed rate. The rate constants follow the Arrhenius relation
    \begin{equation}
        k_i(T_r) = A_i \exp\left[\frac{E_{i,A}}{R} \left(\frac{1}{T_{r,i}} - \frac{1}{T_r}\right)\right],
    \end{equation}
    and the energy balance is
    \begin{equation}
        \dot{T}_r = \frac{UA(T_a - T_r) - F C_{p,A}(T_r - T_0) + V \sum_j (-\Delta H_{r,j})\, \mathcal{R}_j}{V \sum_i C_i C_{p,i} + N_{H_2SO_4} C_{p,H_2SO_4}},
    \end{equation}
    where $\mathcal{R}_j$ is the rate of reaction $j$. The inputs are the feed rate $F \in [0, 270]$ and the jacket temperature $T_a \in [298, 500]$. The objective is the amount of product $C$ at the end of the batch, given as a reward on the final transition only,
    \begin{equation}
        r_t =
        \begin{cases}
            \dfrac{C_{C,T}\, V_T}{(C_C V)_{\max}}, & t = T-1, \\[4pt]
            0, & \text{otherwise,}
        \end{cases}
    \end{equation}
    where $C_{C,T}$ and $V_T$ are reached after applying $\bu_{T-1}$, and $(C_C V)_{\max}$, the largest attainable amount of product, maps the best achievable reward to one. The temperature and volume are constrained to $T_r \le 420$\,K and $V \le 800$\,L. Every time step subtracts the penalty $0.5 \sum_k \max(0, g_k / \bar g_k - 1)$ from the reward, where $g_k$ is a constrained state after the step and $\bar g_k$ its bound. Without the penalty the best policy heats the reactor to about 560\,K, and with it the best policy keeps the temperature near its bound.
    
    \paragraph{Stochasticity.} Each integration step adds multiplicative Gaussian state noise, $\bx_{t+1} \leftarrow \bx_{t+1} + \boldsymbol{\xi}_t \odot \bx_{t+1}$ with $\boldsymbol{\xi}_t \sim \mathcal{N}(\mathbf{0}, \sigma_x^2 I)$ and $\sigma_x = 0.02$ (2\% relative noise). Initial states are $\bx_0 = \bar{\bx}_0 + \boldsymbol{\zeta}$ with $\bar{\bx}_0 = [1.0, 0.0, 0.0, 290.0, 100.0]^\top$ and $\boldsymbol{\zeta} \sim \mathcal{U}([-0.2, 0.2] \times \{0\} \times \{0\} \times [-10, 10] \times [-20, 20])$, with the initial temperature floored at $280$\,K and the initial volume at $50$\,L. Before each time step the pre-exponential factors $A_1$ and $A_2$ are multiplied by independent factors $\exp(s\varepsilon - s^2/2)$ with $\varepsilon \sim \mathcal{N}(0, 1)$ and $s = 0.6$, as when catalyst activity, mixing and mass transfer vary between control intervals. The factors have mean one. Through the Arrhenius rates the resulting noise is small while the batch is cold and large near the temperature bound.
    
    \subsection{Photosynthetic biomass production}
    
    This benchmark models light- and nutrient-driven growth of the cyanobacterium \textit{Arthrospira platensis} and its production of the pigment phycocyanin \citep{bloor2025pc}. The state $\bx = [c_x, c_N, c_q]^\top$ tracks the biomass concentration (g/L), the nitrate concentration (mg/L) and the phycocyanin concentration (g/L), with
    \begin{align}
        \dot{c}_x &= g(I, c_N)\, c_x - u_d\, c_x, \\
        \dot{c}_N &= -Y_{NX}\, g(I, c_N)\, c_x + F_N, \\
        \dot{c}_q &= \frac{k_m I}{I + k_{sq} + I^2/k_{iq}}\, c_x - \frac{k_d\, c_q}{c_N + K_{Nq}},
    \end{align}
    where the light- and nitrate-limited specific growth rate is
    \begin{equation}
        g(I, c_N) = \frac{u_m I}{I + k_s + I^2/k_i} \cdot \frac{c_N}{c_N + k_N},
    \end{equation}
    and the light intensity $I$ and the nitrate feed rate $F_N$ are the inputs. The kinetic parameters are $u_m = 0.0572$, $u_d = 0$, $Y_{NX} = 504.5$, $k_m = 1.6 \times 10^{-4}$, $k_d = 0.281$, $k_{sq} = 23.51$, $K_{Nq} = 16.89$, $k_{iq} = 800$, $k_s = 178.9$, $k_i = 447.1$ and $k_N = 393.1$. The inputs are constrained to $I \in [120, 400]\,\mu\text{E/m}^2\text{/s}$ and $F_N \in [0, 40]$\,mg/L/h, and the states are scaled for the GP by the ranges $c_x \in [0, 20]$\,g/L, $c_N \in [0, 7000]$\,mg/L and $c_q \in [0, 0.15]$\,g/L. The objective is the final biomass, given as a reward on the final transition only,
    \begin{equation}
        r_t =
        \begin{cases}
            \dfrac{c_{x,T}}{c_{x,\max}}, & t = T-1, \\[4pt]
            0, & \text{otherwise,}
        \end{cases}
    \end{equation}
    where $c_{x,T}$ is reached after applying $\bu_{T-1}$, and $c_{x,\max} \approx 17.0$\,g/L is the largest final biomass reached by constant inputs within the bounds (found by a grid search), which maps that value to one. The nitrate concentration is constrained to $c_N \le 800$\,mg/L, which a penalty enforces: every time step subtracts the penalty $0.05 \max(0, c_N / 800 - 1)$ from the reward. Rewards are divided by 10 before learning.
    
    \paragraph{Stochasticity.} Each integration step adds multiplicative Gaussian state noise as above with $\sigma_x = 0.001$ (0.1\% relative noise). Initial states are $\bx_0 = \bar{\bx}_0 + \boldsymbol{\zeta}$ with $\bar{\bx}_0 = [0.1, 20.0, 0.01]^\top$ and $\boldsymbol{\zeta} \sim \mathcal{U}([-0.02, 0.02] \times [-2.0, 2.0] \times [-0.002, 0.002])$, floored at zero componentwise. Before each time step the maximum specific growth rate $u_m$ is multiplied by a factor $\exp(s\varepsilon - s^2/2)$ with $\varepsilon \sim \mathcal{N}(0, 1)$ and $s = 0.6$, as when culture temperature and light vary between control intervals. The resulting noise grows with the biomass density and the growth rate, so it is small early in the batch and large near its end. The horizon is $T = 8$ time steps over a simulated window of 240 hours.
    
    \subsection{Inverted pendulum}
    
    A pendulum of mass $m = 1$ and length $l = 1$ under gravity $g = 9.8$ has angle $\theta$, zero when upright, and angular velocity $\dot\theta$. The input is a torque $u \in [-2, 2]$, and
    \begin{equation}
    \ddot\theta = \frac{3g}{2l} \sin\theta + \frac{3}{m l^2}\, u,
    \end{equation}
    integrated by the explicit Euler method with two substeps per time step of $\Delta t = 0.2$, over $T = 20$ time steps. The controller observes $[\sin\theta, \cos\theta, \dot\theta]^\top$, which avoids the discontinuity of the angle at $\pm\pi$, and $\dot\theta$ is scaled by the bound $8$ for the GP inputs. The pendulum starts hanging down and must be swung up. Each time step costs $0.05\, u_t^2\, \Delta t$, and the final time step is instead rewarded by
    \begin{equation}
    r_{T-1} = -|\theta_T| - 0.1\, |\dot\theta_T|,
    \end{equation}
    with $\theta$ not wrapped. Rewards are divided by 10 before learning.
    
    \paragraph{Stochasticity.} Initial states are $\theta_0 \sim \mathcal{N}(\pi, 0.3^2)$ and $\dot\theta_0 \sim \mathcal{N}(0, 0.3^2)$. After each time step, $\mathcal{N}(0, 0.05^2)$ is added to $\dot\theta$, a random external torque.
    
    \subsection{Target tracking}
    
    A point mass with position $\mathbf{p} = (y, z)$ and unit mass hangs in unit gravity from a spring of unit stiffness attached to an anchor $\mathbf{a} = (a_y, a_z)$, whose rate of change is the input $\bu = \dot{\mathbf{a}} \in [-1, 1]^2$. The state $\bx = [\mathbf{a}, \mathbf{p}, \dot{\mathbf{p}}]^\top \in \mathbb{R}^6$ evolves as
    \begin{equation}
    \dot{\mathbf{a}} = \bu, \qquad \ddot{\mathbf{p}} = -(\mathbf{p} - \mathbf{a}) - \begin{bmatrix} 0 \\ 1 \end{bmatrix},
    \end{equation}
    integrated by the fourth-order Runge--Kutta method with ten substeps per time step of $\Delta t = 1$, over $T = 10$ time steps. Each time step costs $10^{-3} \|\bu_t\|^2 \Delta t$, and the final time step is instead rewarded by
    \begin{equation}
    r_{T-1} = -\|\mathbf{a}_T\|^2 - \|\mathbf{p}_T - \mathbf{p}^*\|^2, \qquad \mathbf{p}^* = (2, 2),
    \end{equation}
    so the mass must reach the target with the anchor back at the origin. The GP inputs are scaled by the bounds $\pm 5$ on every state component. Rewards are divided by 10 before learning.
    
    \paragraph{Stochasticity.} Initial states are $\bx_0 \sim \mathcal{N}(\mathbf{0}, \operatorname{diag}(0.2, 0.2, 0.3, 0.3, 0.1, 0.1)^2)$. After each time step, $\mathcal{N}(0, 0.03^2)$ is added to every state component.
    

    \section{Synthetic MDPs}\label{app:toy}

    Both synthetic MDPs of Section~\ref{sec:results_toy} have a state $x \in [-1, 1]$, an input $u \in [-1, 1]$, horizon $T = 3$, transitions $x' = \operatorname{clip}(x + 0.35\,u + b(x)\,e, -1, 1)$ with $e \sim \mathcal{N}(0, 1)$, a stage reward $-0.05\,u^2$ and a terminal reward $r_T(x')$. In A, $b(x) = 0.02 + 0.25\,\operatorname{logistic}(-(x + 0.2)/0.1)$ and $r_T(x) = 1 - x^2$. In B, $b(x) = 0.01$ and $r_T(x) = \operatorname{logistic}((x - 0.3)/0.03)$. Each dataset holds 150 transitions per time step with states and inputs drawn uniformly. Every method fits HGPQ's GP of Appendix~\ref{app:fitting} backwards from the terminal reward, with $V_{t+1}(x')$ the maximum of $\mu_{t+1}(x', \cdot)$ over 51 inputs, and differs only in the target variances: the floor $\sigma_r^2 = 10^{-4}$, one noise level per time step fitted by marginal likelihood, or HGPQ's $\sigma_r^2 + \omega_t$. The optimal values come from dynamic programming on a grid of 201 states and 51 inputs. The return gap is the difference between the optimal mean return and that of the greedy policy, simulated from 41 start states with twenty noise draws each.

    \paragraph{Results.}\label{app:toy_results} Table~\ref{tab:toy} gives the return gap and the mean absolute error of the value learned at time step 1 on a grid of states, averaged over the twenty datasets, for $x \ge 0$ and for $x < 0$, which in B contains the step of $V^*_1$. HGPQ's gap is smaller than that of one noise level by $0.003 \pm 0.003$ in A and $0.012 \pm 0.010$ in B (paired over the twenty datasets, 95\% intervals), and smaller on fifteen of the twenty datasets in each.

    \begin{table}[h]
    \centering
    \caption{Synthetic MDPs: return gap of the greedy policy and mean absolute value error at time step 1, averaged over twenty datasets.}
    \label{tab:toy}
    \footnotesize
    \renewcommand{\arraystretch}{1.1}
    \begin{tabular}{lcccccc}
    \toprule
     & \multicolumn{3}{c}{\textbf{A: state-dependent noise}} & \multicolumn{3}{c}{\textbf{B: step in the value}} \\
    \cmidrule(lr){2-4} \cmidrule(lr){5-7}
    \textbf{Target variance} & gap & error, $x \ge 0$ & error, $x < 0$ & gap & error, $x \ge 0$ & error, $x < 0$ \\
    \midrule
    Floor $\sigma_r^2$ & 0.229 & 0.147 & 1.26 & 0.041 & 0.163 & 0.110 \\
    One noise level per time step & 0.008 & 0.011 & 0.035 & 0.035 & 0.105 & \textbf{0.098} \\
    HGPQ & \textbf{0.006} & \textbf{0.006} & \textbf{0.029} & \textbf{0.023} & \textbf{0.023} & 0.121 \\
    \bottomrule
    \end{tabular}
    \end{table}

    \section{Baselines, protocol and hyperparameters}\label{app:protocol}\label{app:hparams}
    
    \paragraph{Baselines.} GPQ-CLR \citep{bloor2025gaussian} and GPQ-SLR, as published, fit one GP over the state, the input and the time step, with one GP per time step only on the LMDP, and use Thompson~sampling without the uniform component. They store a value for each visited state--input pair and move it towards its target with a constant or a $1/N_{\mathcal{D}}$ learning rate. KOVI \citep{yang2020provably} fits a kernel ridge regression at each time step, backwards in time, with a single noise level, and acts on, and bootstraps from, the posterior mean plus a multiple $c$ of the posterior standard deviation. GPTD, in its SARSA form \citep{engel2005reinforcement}, keeps one GP over the Q-function, observes each reward through the difference of the values of successive state--input pairs, acts on the posterior mean and explores with an annealed $\epsilon$-greedy rule. KTD \citep{geist2010kalman} is a Kalman filter on random Fourier features. The deep RL baselines are DQN \citep{mnih2015human}, Ensemble-DQN \citep{osband2016deep}, DDPG \citep{lillicrap_continuous_2019}, TD3 \citep{fujimoto_addressing_2018} and SAC \citep{haarnoja2018soft}.

    \paragraph{Hyperparameters.} HGPQ has no learning rate. Its tuned parameters are the reward noise $\sigma_r$ and, for the constant choice of the target variance, the scale $s$ in the constant variance $\sigma_r^2 + s\,\kappa_{\max}$. Below are the fixed settings of every method, the tuning grids and the selected values (Table~\ref{tab:settings}). Every method is tuned on ten seeds (100 to 109) that are not used for evaluation, by the mean final return, with ties within one standard error broken by the area under the learning curve. When the selected value lies at an edge of its grid, the grid is extended beyond that edge by two values at the ratio of the grid and the selection is repeated.
    
    \paragraph{HGPQ.} The GPs are fitted as in Appendix~\ref{app:implementation}, with the fixed settings of Table~\ref{tab:hgpq_fixed}. On the LMDP the GP input also contains the features $|p_t|$ and $\operatorname{sign}(p_t)\,\bu_t$. In the constant variance $\sigma_r^2 + s\,\kappa_{\max}$, $\kappa_{\max} = \sup_{\bz} \kappa(\bz, \bz) = \sigma_f^2$ is the prior variance of the kernel refitted at time step $t+1$, in the units of the targets. Table~\ref{tab:settings} gives the per-step reward bounds $[r_{\min}, r_{\max}]$ that clip the targets, in the scaled units of Appendix~\ref{app:models}. We tune $\sigma_r \in \{10^{-4}, 10^{-3}, 10^{-2}, 0.1\}$ and, for the constant target variance, $s \in \{0, 10^{-4}, 5 \times 10^{-4}, 2 \times 10^{-3}, 10^{-2}, 0.1, 1\}$ jointly. Appendix~\ref{app:floor} reports the sensitivity to $\sigma_r$ and to $\epsilon$.

    \begin{table}[h]
    \centering
    \caption{Fixed settings of HGPQ.}
    \label{tab:hgpq_fixed}
    \footnotesize
    \renewcommand{\arraystretch}{1.1}
    \begin{tabular}{@{}>{\raggedright\arraybackslash}p{4.2cm}>{\raggedright\arraybackslash}p{9.4cm}@{}}
    \toprule
    \textbf{Setting} & \textbf{Value} \\
    \midrule
    Initial episodes $N_0$ & 5 (LMDP), 10 (pendulum, target tracking), 20 (reactor, biomass production) \\
    Uniform exploration probability & $\epsilon = 0.1$ \\
    Thompson draws & pathwise conditioning with 1024 random Fourier features \\
    Candidate inputs & 500: the lower corner, the upper corner and the centre of the input box, then Sobol points \\
    Local maximisation & L-BFGS-B from the 5 best candidates, at most 50 iterations each \\
    Candidate inputs on the LMDP & $\{-1, 1\}$, since the reward is linear in the input \\
    \bottomrule
    \end{tabular}
    \end{table}
    
    \paragraph{GPQ-SLR and GPQ-CLR.} A GP with HGPQ's Mat\'ern-5/2 kernel, refit by marginal likelihood, on a table of stored values. Each Bellman target moves its stored value with learning rate $1/N_{\mathcal{D}}$ (GPQ-SLR, with $N_{\mathcal{D}}$ the number of stored values) or a constant $\alpha$ (GPQ-CLR). Both use HGPQ's initial episodes and candidate inputs, and Thompson~sampling without the uniform component. GPQ-SLR has no free parameter. For GPQ-CLR we tune $\alpha \in \{0.01, 0.03, 0.1, 0.3\}$, capped at 1.
    
    \paragraph{KOVI.} We implement KOVI as in the kernel setting of \citet{yang2020provably}, with one kernel ridge regression per time step, using a squared-exponential kernel of unit variance on inputs scaled to $[-1, 1]$, refitted by backward induction after every episode, and a bonus proportional to the posterior standard deviation. Every ten episodes, the lengthscale of each time step is chosen by marginal likelihood. KOVI acts on, and bootstraps from, the posterior mean plus $c$ posterior standard deviations, clipped to the range of possible returns. Its regulariser $\lambda$ is the one selected, over $\{0.01, 0.1, 1, 10\}$, for the same regression without the bonus, with a fixed unit lengthscale and annealed $\epsilon$-greedy exploration, and we tune $c \in \{0.01, 0.03, 0.1, 0.3, 1\}$. Bootstrapped values maximise over 100 candidate inputs, and the behaviour policy over 500, refined by L-BFGS-B.
    
    \paragraph{GPTD.} We implement GPTD in its SARSA form (GPSARSA) as \citet{engel2005reinforcement} describe it. A GP models the Q-function, and along each episode every reward is observed through $r_t = Q_t(\bz_t) - Q_{t+1}(\bz_{t+1}) + N_t$, with $Q_T :\equiv 0$ and the noise of their model for stochastic transitions, whose covariance over an episode is $\sigma^2 H H^\top$ for the temporal-difference matrix $H$. The observation is linear in $Q$, so the posterior is exact under this model. For a complete episode $H$ is invertible, and the posterior mean is GP regression of the returns-to-go on the visited pairs with noise variance $\sigma^2$, which we compute exactly after every episode. Each time step has an independent GP prior on inputs scaled to $[-1, 1]$. GPTD acts on the posterior mean, maximised as for KOVI, and explores by $\epsilon$-greedy with $\epsilon = 10/(10 + k)$ in episode $k$. We tune $\sigma^2 \in \{10^{-3}, 10^{-2}, 0.1, 1\}$, extended to $10$ and $100$, or to $10^{-4}$ and $10^{-5}$, where the selected value lay at an edge.
    
    \paragraph{KTD.} A linear Kalman filter at each time step on 500 random Fourier features, with prior covariance $10\, I$, observation variance $0.1$ and process noise $10^{-3} I$. We tune the lengthscale of the features over $\{0.5, 1, 2\}$, extended at an edge as above. 

    \paragraph{KFQL.} Kalman filter Q-learning \citep{tripp2013approximate} on KTD's features and lengthscales, without process noise. The observation variance of a target is a constant $\varepsilon_0$ plus $\gamma^2 \boldsymbol{\phi}(\bx', \bu^*)^\top P_{t+1} \boldsymbol{\phi}(\bx', \bu^*)$, the variance of the successor's value at its greedy input under the next time step's filter. This is the form of variance that Section~\ref{sec:noise} argues against, here on a parametric model. We tune $\varepsilon_0$ over $\{10^{-4}, 10^{-3}, \ldots, 1\}$, extended at the edges.
    
    \paragraph{Deep RL.} Table~\ref{tab:deeprl} lists the settings of the deep RL baselines. Observations and inputs are normalised to $[-1, 1]$ with their bounds. DQN maximises over continuous inputs by L-BFGS-B on $[-1, 1]^{d_u}$ and Ensemble-DQN by projected gradient ascent with three restarts. Ensemble-DQN keeps each transition for each critic with probability $0.8$, explores by the upper-confidence rule over the ensemble with coefficient $\beta$ and is evaluated with the ensemble mean. On the reactor and biomass production SAC uses a target entropy scale of $0.5$ and multiplies the reward by 50 to compensate for the sparse terminal reward.

    \begin{table}[h]
    \centering
    \caption{Settings of the deep RL baselines. Learning-rate grids are multiplied by $\{1/3, 1, 3\}$ and extended at an edge as above.}
    \label{tab:deeprl}
    \footnotesize
    \renewcommand{\arraystretch}{1.1}
    \setlength{\tabcolsep}{3pt}
    \begin{tabular}{@{}>{\raggedright\arraybackslash}p{2.3cm}>{\raggedright\arraybackslash}p{2.2cm}>{\raggedright\arraybackslash}p{2.5cm}>{\raggedright\arraybackslash}p{2.3cm}>{\raggedright\arraybackslash}p{2.4cm}>{\raggedright\arraybackslash}p{2.2cm}@{}}
    \toprule
     & \textbf{DQN} & \textbf{Ensemble-DQN} & \textbf{DDPG} & \textbf{TD3} & \textbf{SAC} \\
    \midrule
    Networks & $128 \times 128$ & five $128 \times 128$ critics & $128 \times 128$ actor and critic & $256 \times 256$ actor and critic & $256 \times 256$ actor and critic \\
    Batch size & 64 & 64 & 64 & 256 & 256 \\
    Replay capacity & $5 \times 10^4$ & $5 \times 10^4$ & $5 \times 10^4$ & $10^6$ & $10^6$ \\
    Target updates & hard, every 100 steps & hard, every 200 steps & soft, $\tau = 5 \times 10^{-3}$ & soft, $\tau = 5 \times 10^{-3}$ & soft, $\tau = 5 \times 10^{-3}$ \\
    Exploration & $\epsilon$-greedy, $\epsilon$ from 1.0 to 0.05, decay 0.99 per episode & upper confidence over the ensemble, $\beta$ & Gaussian noise 0.1 & Gaussian noise 0.1 & entropy, target scale 1.0 \\
    Other & gradient norm clipped at 1 & & & target noise 0.2 clipped at 0.5, policy update every second step, learning after 1000 steps & learning after 200 steps \\
    Tuned & learning rate $3 \times 10^{-4}$ & learning rate $10^{-3}$, $\beta \in \{0.5, 1, 2\}$ & actor and critic learning rates $(10^{-4}, 10^{-3})$ & learning rate $3 \times 10^{-4}$ & learning rate $3 \times 10^{-4}$ \\
    \bottomrule
    \end{tabular}
    \end{table}

    \begin{table}[h]
    \centering
    \small
    \caption{Reward bounds of HGPQ and the selected value of every tuned parameter (rewards in the scaled units of Appendix~\ref{app:models}). DDPG's critic learning rate is ten times its actor learning rate.}
    \label{tab:settings}
    \renewcommand{\arraystretch}{1.1}
    \footnotesize
    \setlength{\tabcolsep}{4pt}
    \begin{tabular}{llccccc}
    \toprule
    \textbf{Method} & \textbf{Parameter} & \textbf{LMDP} & \textbf{Reactor} & \textbf{Biomass} & \textbf{Pendulum} & \textbf{Target} \\
    \midrule
    HGPQ & $[r_{\min}, r_{\max}]$ & $[-0.1, 0.1]$ & $[-0.5, 1]$ & $[-0.03, 0.1]$ & $[-6, 0]$ & $[-65, 0]$ \\
         & $\sigma_r$ & $10^{-2}$ & $10^{-2}$ & $10^{-3}$ & $10^{-2}$ & $10^{-4}$ \\
         & $s$ (constant) & $5 \times 10^{-4}$ & $10^{-2}$ & $10^{-2}$ & $2 \times 10^{-3}$ & $5 \times 10^{-4}$ \\
    GPQ-CLR & $\alpha$ & $0.3$ & $10^{-2}$ & $0.3$ & $1.1 \times 10^{-3}$ & $1.1 \times 10^{-3}$ \\
    KOVI & $c$ & $0.03$ & $0.3$ & $0.03$ & $0.1$ & $0.1$ \\
    KOVI & $\lambda$ & $10^{-2}$ & $10^{-2}$ & $10^{-2}$ & $10^{-2}$ & $10^{-3}$ \\
    GPTD & $\sigma^2$ & $10$ & $1$ & $1$ & $0.1$ & $10^{-4}$ \\
    KTD & lengthscale & $4$ & $2$ & $2$ & $0.125$ & $2$ \\
    KFQL & $\varepsilon_0$ & $0.1$ & $10^{-3}$ & $10^{-6}$ & $1$ & $10^{-4}$ \\
    DQN & learning rate & $3 \times 10^{-4}$ & $8.1 \times 10^{-3}$ & $8.1 \times 10^{-3}$ & $3 \times 10^{-4}$ & $2.7 \times 10^{-3}$ \\
    Ensemble-DQN & learning rate & $9 \times 10^{-3}$ & $3 \times 10^{-3}$ & $3 \times 10^{-3}$ & $3 \times 10^{-3}$ & $3 \times 10^{-3}$ \\
                 & $\beta$ & $2$ & $1$ & $1$ & $1$ & $1$ \\
    DDPG & actor learning rate & $3 \times 10^{-4}$ & $3 \times 10^{-4}$ & $9 \times 10^{-4}$ & $3 \times 10^{-4}$ & $3 \times 10^{-4}$ \\
    TD3 & learning rate & $10^{-4}$ & $10^{-4}$ & $9 \times 10^{-4}$ & $2.7 \times 10^{-3}$ & $10^{-4}$ \\
    SAC & learning rate & $8.1 \times 10^{-3}$ & $8.1 \times 10^{-3}$ & $2.4 \times 10^{-2}$ & $8.1 \times 10^{-3}$ & $9 \times 10^{-4}$ \\
    \bottomrule
    \end{tabular}
    \end{table}
    
    \section{Sparse HGPQ}\label{app:scaling}
    
    This appendix introduces Sparse HGPQ, which replaces each exact GP of HGPQ by a sparse GP. It shows that Sparse HGPQ keeps the cost linear in the number of episodes without changing the algorithm, and that on the pendulum and target tracking it remains more sample-efficient than the deep RL methods when they need hundreds of episodes.

    \paragraph{Sparse HGPQ.} An exact refit after $n$ episodes costs $O(n^3)$ per time step, so for budgets of hundreds of episodes Sparse HGPQ replaces the GP of each time step by the sparse variational GP of \citet{titsias2009variational} with $m$ inducing inputs, here with the per-target variances $\lambda_{t,i} = \sigma_r^2 + \omega_t(\bz_i)$ of Section~\ref{sec:noise} in place of a single noise level. The posterior mean becomes
    \begin{equation}\label{eq:sparse_mean}
    \mu_t(\bz) = \bkappa_m(\bz)^\top A^{-1} \sum_{i=1}^{n} \frac{\bkappa_m(\bz_i)\, y_i}{\lambda_{t,i}}, \qquad A = K_{mm} + \sum_{i=1}^{n} \frac{\bkappa_m(\bz_i)\, \bkappa_m(\bz_i)^\top}{\lambda_{t,i}},
    \end{equation}
    where $\bkappa_m(\bz)$ holds the kernel between $\bz$ and the inducing inputs. Every target enters, weighted by $1/\lambda_{t,i}$, but only the $m \times m$ matrix $A$ is inverted, so a refit costs $O(nm^2)$, and the exact posterior is recovered when the inducing inputs are the training inputs. The kernel hyperparameters maximise the collapsed lower bound of \citet{titsias2009variational} with these variances, which are estimated from leave-one-out residuals as for the exact GP (Appendix~\ref{app:fitting}), and the inducing inputs are located by a pivoted Cholesky decomposition of the prior kernel matrix, which \citet{burt2020convergence} show places them close to optimally. Everything else in Algorithm~\ref{alg:hgpq} is unchanged, and we use $m = 256$ unless stated otherwise.

    \paragraph{Benchmarking.} We make the benchmarks harder by multiplying the process noise of the pendulum, target tracking and the reactor by three and by ten, by controlling $k$ independent pendulums at once, which enlarges the state--input pair $\bz$ to dimension $d = 4k$, and by lengthening the horizon of the pendulum over the same physical time, with $T$ from 10 to 80 and time step $4/T$. Every method keeps its settings of Appendix~\ref{app:hparams} without tuning them again. We compare Sparse HGPQ with SAC, TD3, DDPG and Ensemble-DQN only, showing exact HGPQ for reference. Sparse HGPQ runs for 600 episodes and the deep RL methods for 2000. Each setting uses ten seeds.
    
\begin{figure}[t]
        \centering
        \includegraphics[width=0.85\linewidth]{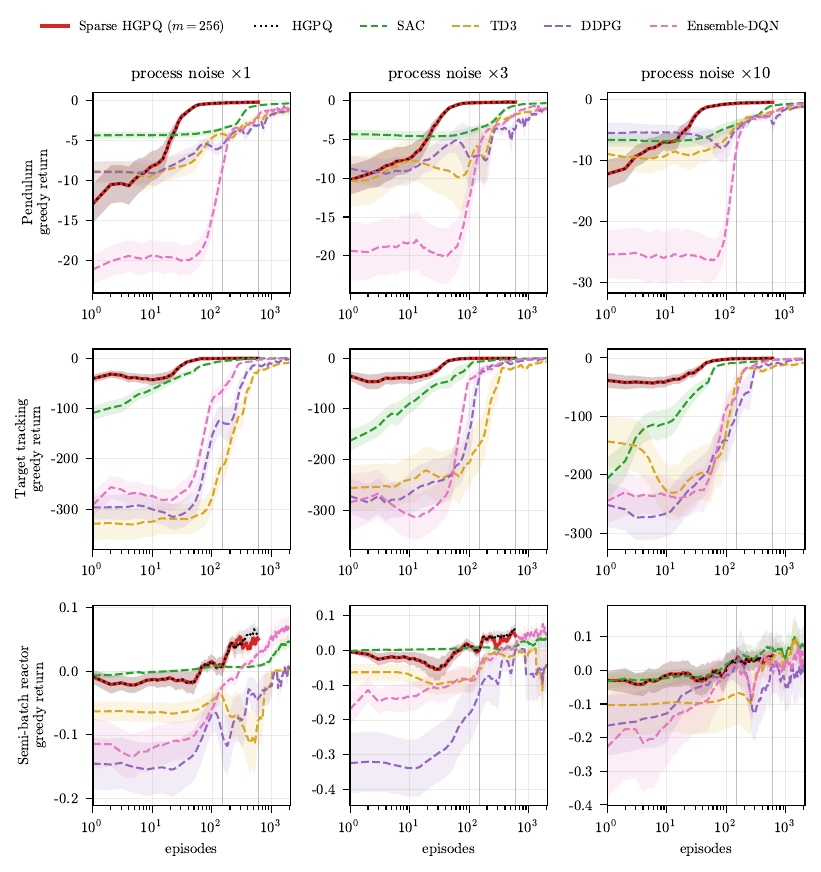}
        \caption{Mean greedy return against episodes with one, three and ten times the process noise. Sparse HGPQ uses 256 inducing points. HGPQ, with exact GPs, is shown for comparison. Both run for 600 episodes and the deep RL methods for 2000. Bands are 95\% confidence intervals over ten seeds.}
        \label{fig:scaling_noise}
    \end{figure}

\begin{figure}[t]
        \centering
        \includegraphics[width=\linewidth]{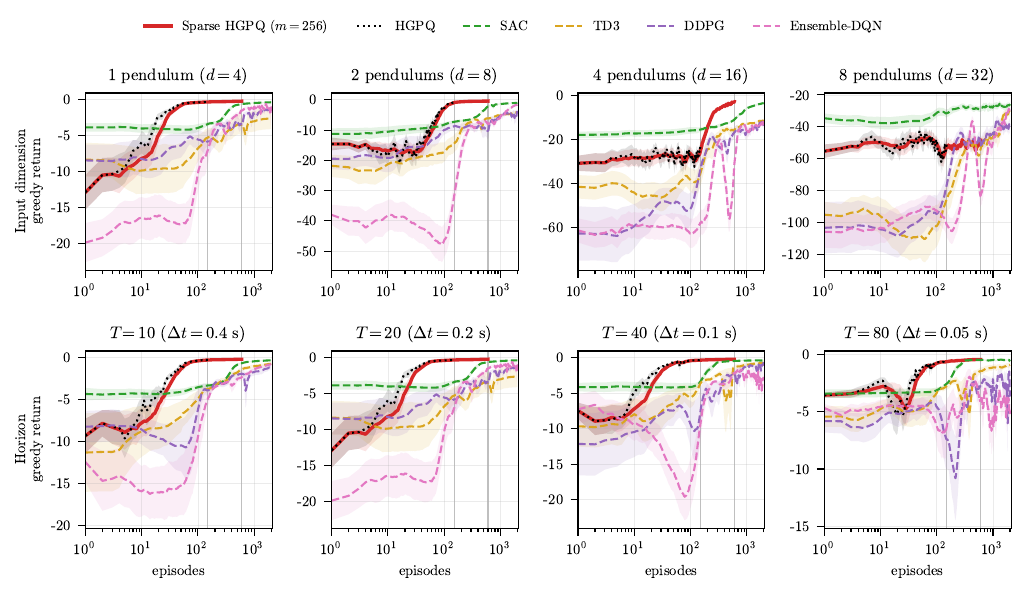}
        \caption{Mean greedy return against episodes. Top: $k$ independent pendulums, with $d = d_x + d_u = 4k$, return summed over the pendulums. Bottom: the pendulum at horizons $T = 10$ to $80$, the same physical task with time step $4/T$. Sparse HGPQ uses 256 inducing points and runs for 600 episodes. HGPQ, with exact GPs, is shown for comparison over 150 episodes. Bands are 95\% confidence intervals over ten seeds.}
        \label{fig:scaling_dim_horizon}
    \end{figure}

    Figures~\ref{fig:scaling_noise} and~\ref{fig:scaling_dim_horizon} show the learning curves. Measured by the number of episodes needed to reach 90\% of the way from the median initial return to the best final return, On the pendulum and target tracking Sparse HGPQ is the fastest method in every setting except target tracking at ten times the process noise, where SAC needs 31 episodes and Sparse HGPQ 35, and its advantage holds as the noise and the horizon grow. On the reactor, whose fluctuating rates make the return vary widely between episodes, Sparse HGPQ is level with or above the deep RL methods over its 600 episodes, but the best final returns come from deep RL after 2000 episodes. Within 600 episodes it reaches the level only at three times the noise, after about 500 episodes against about 800 for Ensemble-DQN, and at one and ten times the noise Ensemble-DQN, DDPG or SAC reach it within their 2000 episodes. The advantage narrows as the dimension $d$ of the state--input pair grows, and with four pendulums Sparse HGPQ needs about 300 episodes and SAC about 900.
    
\begin{figure}[t]
        \centering
        \includegraphics[width=0.9\linewidth]{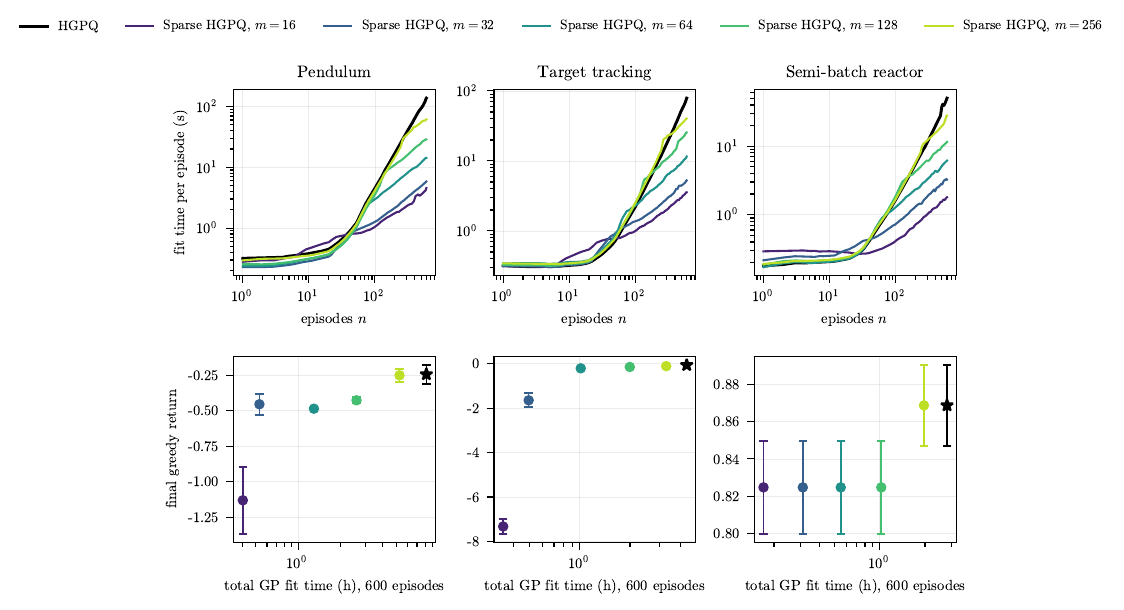}
        \caption{Top: time per episode spent fitting the GPs of HGPQ and of Sparse HGPQ with $m$ inducing points. Bottom: final return after 600 episodes against the total fitting time, mean $\pm$ standard error over three to ten seeds. The reactor runs use its version without the constraint penalty and the rate fluctuations.}
        \label{fig:scaling_compute}
    \end{figure}

    \FloatBarrier
    \paragraph{Cost of the GP fit.} Figure~\ref{fig:scaling_compute} shows the time per episode spent fitting the GPs over 600 episodes, for HGPQ and for Sparse HGPQ with 16 to 256 inducing points, both with a constant target variance $\sigma_r^2 + s\,\kappa_{\max}$, which leaves out the cost of the heteroscedastic estimate. The cost of HGPQ grows faster than linearly in the number of episodes, and that of Sparse HGPQ grows linearly. With 128 or 256 inducing points Sparse HGPQ reaches a final return close to that of HGPQ, and with fewer its return falls on the pendulum and on target tracking.
    
    \FloatBarrier
    \section{Ablations}\label{app:ablation}

    This appendix tests the use of the next time step's uncertainty in the target variances, the link between the variances and their targets, the choice of the noise estimator, and the sensitivity to the floor $\sigma_r$ and to exploration.

    \paragraph{Posterior variance of the next time step.}\label{app:ablation_next} Table~\ref{tab:factorial} compares HGPQ with two homoscedastic variances and with three variances that contain the posterior variance $\sigma^2_{t+1}(\bx'_i, \bu^*_i)$ of the next GP: HGPQ's variance with $\sigma^2_{t+1}$ added, the floor plus $\sigma^2_{t+1}$, and a tuned constant plus $\sigma^2_{t+1}$, all with HGPQ's GP, exploration and settings and on the same seeds. Adding $\sigma^2_{t+1}$ to HGPQ's variance changes the final return by at most 5\%, and significantly only on target tracking, where it lowers the final cost from 0.097 to 0.069. It lowers the mean return of the first fifty episodes on the pendulum and target tracking (paired $t$-test over seeds, $p < 0.05$) and leaves it unchanged within the noise on the reactor and biomass production. The floor plus $\sigma^2_{t+1}$, which gives each target the uncertainty of the next estimate but no estimate of its own error, is worse on the reactor and biomass production. A tuned constant plus $\sigma^2_{t+1}$, with $s$ tuned on the grid of Appendix~\ref{app:hparams}, matches HGPQ's final return on the pendulum and target tracking but not on the reactor and biomass production, and starts more slowly on the pendulum, the reactor and biomass production. On target tracking every variance without HGPQ's estimate starts faster, but the homoscedastic ones end with a higher cost.

    \begin{table}[t]
    \centering
    \caption{Choices of target variance on HGPQ's GP: final return after 150 episodes (top) and mean return over the first fifty episodes (bottom), in the units of each benchmark, mean and 95\% interval over seeds. Bold: the best mean of each row and every entry whose interval overlaps its interval. Thirty seeds (fifty on the pendulum and target tracking), tuned constant thirty. On the pendulum and the reactor the tuned scale of the last column is $s = 0$, so it equals the floor plus $\sigma^2_{t+1}$.}
    \label{tab:factorial}
    \scriptsize
    \renewcommand{\arraystretch}{1.05}\setlength{\tabcolsep}{2pt}
    \begin{tabular}{lcccccc}
    \toprule
    & & \multicolumn{2}{c}{\textbf{Homoscedastic}} & \multicolumn{3}{c}{\textbf{With the next GP's variance}} \\
    \cmidrule(lr){3-4} \cmidrule(lr){5-7}
    \textbf{Setting} & \textbf{HGPQ} & $\sigma_r^2$ & $\sigma_r^2 + s\,\kappa_{\max}$ & HGPQ $+\,\sigma^2_{t+1}$ & $\sigma_r^2 + \sigma^2_{t+1}$ & \shortstack{$\sigma_r^2 + s\,\kappa_{\max}$\\$+\,\sigma^2_{t+1}$} \\
    \midrule
    \multicolumn{7}{l}{\textit{Final return after 150 episodes}} \\
    Reactor & $\mathbf{0.048\,{\pm}\,0.014}$ & $0.004\,{\pm}\,0.022$ & $\mathbf{0.020\,{\pm}\,0.019}$ & $\mathbf{0.046\,{\pm}\,0.013}$ & $\mathbf{0.026\,{\pm}\,0.015}$ & $\mathbf{0.026\,{\pm}\,0.015}$ \\
    Biomass & $\mathbf{0.137\,{\pm}\,0.017}$ & $-0.356\,{\pm}\,0.097$ & $0.081\,{\pm}\,0.038$ & $\mathbf{0.144\,{\pm}\,0.013}$ & $-0.049\,{\pm}\,0.066$ & $0.111\,{\pm}\,0.019$ \\
    Pendulum & $\mathbf{-0.26\,{\pm}\,0.03}$ & $-1.81\,{\pm}\,0.40$ & $-1.11\,{\pm}\,0.68$ & $\mathbf{-0.25\,{\pm}\,0.03}$ & $\mathbf{-0.29\,{\pm}\,0.07}$ & $\mathbf{-0.29\,{\pm}\,0.07}$ \\
    Target & $-0.097\,{\pm}\,0.013$ & $-0.271\,{\pm}\,0.031$ & $-0.224\,{\pm}\,0.049$ & $\mathbf{-0.069\,{\pm}\,0.013}$ & $\mathbf{-0.074\,{\pm}\,0.020}$ & $\mathbf{-0.097\,{\pm}\,0.016}$ \\
    \midrule
    \multicolumn{7}{l}{\textit{Mean return over the first fifty episodes}} \\
    Reactor & $\mathbf{-0.013\,{\pm}\,0.006}$ & $\mathbf{-0.021\,{\pm}\,0.011}$ & $\mathbf{-0.019\,{\pm}\,0.012}$ & $\mathbf{-0.013\,{\pm}\,0.005}$ & $\mathbf{-0.027\,{\pm}\,0.010}$ & $\mathbf{-0.027\,{\pm}\,0.010}$ \\
    Biomass & $\mathbf{0.046\,{\pm}\,0.014}$ & $-0.215\,{\pm}\,0.084$ & $-0.087\,{\pm}\,0.055$ & $\mathbf{0.039\,{\pm}\,0.012}$ & $-0.127\,{\pm}\,0.053$ & $-0.012\,{\pm}\,0.023$ \\
    Pendulum & $\mathbf{-3.16\,{\pm}\,0.28}$ & $-4.58\,{\pm}\,0.41$ & $-4.64\,{\pm}\,0.54$ & $-3.96\,{\pm}\,0.33$ & $-4.21\,{\pm}\,0.53$ & $-4.21\,{\pm}\,0.53$ \\
    Target & $-19.4\,{\pm}\,1.3$ & $\mathbf{-12.5\,{\pm}\,0.8}$ & $\mathbf{-12.9\,{\pm}\,1.3}$ & $-21.4\,{\pm}\,1.5$ & $\mathbf{-13.3\,{\pm}\,0.9}$ & $\mathbf{-13.7\,{\pm}\,0.9}$ \\
    \bottomrule
    \end{tabular}
    \end{table}

    \paragraph{Variances moved between targets.}\label{app:matching} Two controls keep HGPQ's variances but detach them from their targets, on twenty seeds paired with HGPQ, with the predictions recorded before the runs. Permuting the variances at random among the targets of each time step, which keeps their values, multiplies the final cost by 6.5 on the pendulum and by 3.2 on target tracking. Giving every target of a time step the geometric mean of its variances does less harm on both, and its effect on the pendulum is not significant.

    \paragraph{Choice of the noise estimator.}\label{app:noise_estimator}
    HGPQ estimates $\omega_t$ with a heteroscedastic GP. A simpler choice is a single level $\omega_t$ per time step, fitted by marginal likelihood together with the kernel. With a few hundred targets per time step and no repeated inputs, the kernel can then pass close to every target, and the fitted level falls towards its lower bound.

    Table~\ref{tab:estimators} compares the heteroscedastic estimate with this single level and with four alternatives on the reactor and biomass production, each with HGPQ's rule $\lambda_{t,i} = \sigma_r^2 + \omega_t(\bz_i)$ and settings. Every alternative lowers the final return on biomass production, and the single level, leave-one-out cross-validation and the full covariance also lower it on the reactor. The heteroscedastic estimate is the best on both benchmarks.

    \begin{table}[h]
    \centering
    \caption{Change of the final return after 150 episodes relative to the heteroscedastic estimate of Section~\ref{sec:noise}, paired over seeds, mean and 95\% interval, in \% of HGPQ's mean.}
    \label{tab:estimators}
    \small
    \begin{tabular}{>{\raggedright\arraybackslash}p{7.4cm}cc}
    \toprule
    \textbf{Estimator of $\omega_t$} & \textbf{Reactor} & \textbf{Biomass} \\
    \midrule
    Single level, marginal likelihood & $-25 \pm 23$ & $-12 \pm 9$ \\
    Single level, log-normal prior & $-4 \pm 29$ & $-11 \pm 6$ \\
    Single level, leave-one-out innovations \citep{mehra1970identification} & $-20 \pm 33$ & $-19 \pm 7$ \\
    Single level, leave-one-out predictive probability for every hyperparameter \citep{bachoc2013cross} & $-42 \pm 32$ & $-10 \pm 7$ \\
    Single level, with the full covariance $S_t$ of the next GP (Appendix~\ref{app:proof_lambda}) & $-105 \pm 37$ & $-74 \pm 12$ \\
    \bottomrule
    \end{tabular}
    \end{table}

    \begin{figure}[t]
        \centering
        \includegraphics[width=\linewidth]{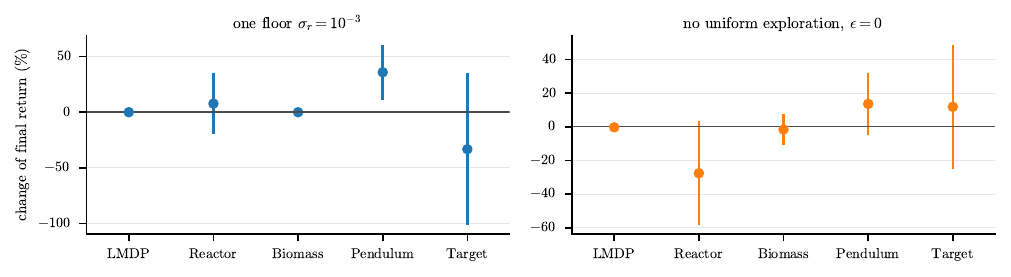}
        \vspace{-4mm}
        \caption{Change of HGPQ's final return after 150 episodes, paired by seed and in \% of HGPQ's mean, with one floor $\sigma_r = 10^{-3}$ on every benchmark (left) and without uniform exploration (right), mean and 95\% interval. On biomass production the tuned floor is already $10^{-3}$.}
        \label{fig:floor_eps}
    \end{figure}

    \paragraph{Floor and exploration.}\label{app:floor} Figure~\ref{fig:floor_eps} shows the sensitivity to the floor and to exploration. With HGPQ's variances, one floor $\sigma_r = 10^{-3}$ on every benchmark instead of the tuned values does not significantly change the final return on the reactor or the LMDP, lowers the final cost of the pendulum by a third, and leaves that of target tracking unchanged within its interval (paired over seeds). On biomass production the tuned value is already $10^{-3}$. Setting $\epsilon = 0$ does not significantly lower the final return on any benchmark, so $\epsilon = 0.1$ is a safeguard for the coverage that Theorem~\ref{thm:main} needs rather than a tuned value.

    \FloatBarrier
\section{Additional results}\label{app:extra_results}
    
\begin{figure}[t]
        \centering
        \includegraphics[width=\linewidth]{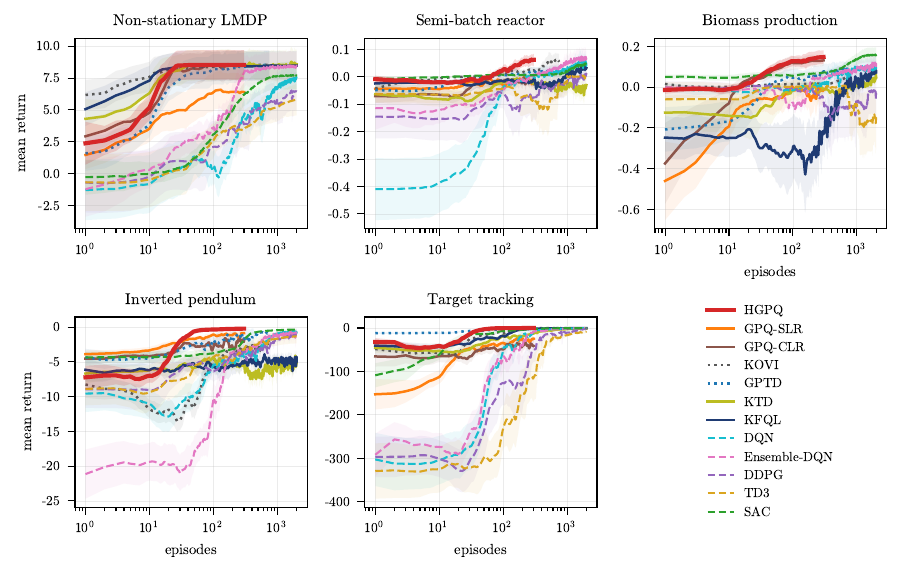}
        \caption{Mean return of the greedy policy against episodes on each benchmark, with 95\% confidence intervals over thirty seeds. Each seed's curve is averaged over the last ten episodes, as in Figure~\ref{fig:ecdf}.}
        \label{fig:curves_raw}
    \end{figure}

    \paragraph{Learning curves.} Figure~\ref{fig:curves_raw} shows the mean return against the number of episodes on each benchmark, in the units of each benchmark. Each curve ends at the end of its method's budget.

 For $x \ge 0$ they are 0.006 for HGPQ, 0.011 with one noise level and 0.147 with the floor alone in A, and 0.023, 0.105 and 0.163 in B. For $x < 0$, where the step of $V^*_1$ lies in B, they are 0.029, 0.035 and 1.26 in A, and 0.121, 0.098 and 0.110 in B.

    \end{document}